%% file: main.tex
\documentclass{article}

\PassOptionsToPackage{numbers,sort,compress}{natbib}
\PassOptionsToPackage{table,dvipsnames}{xcolor}
\usepackage[preprint]{neurips_2026}
\input{headings/packages}

\input{headings/macros}

\title{\textsc{Pace}: Prune-And-Compress Ensemble Models}

\author{%
  Fabian Akkerman\thanks{Authors with equal contributions.}\\
  Industrial Engineering and Management Science\\
  University of Twente, Netherlands\\
  \texttt{f.r.akkerman@utwente.nl}\\
  \And
  Julien Ferry${}^{*}$\\
  SCALE-AI Chair and CIRRELT \\
  Polytechnique Montréal, Canada \\
  \texttt{julien.ferry@polymtl.ca} \\
  \AND
  Théo Guyard${}^{*}$\\
  \phantom{aa}SCALE-AI Chair and CIRRELT\phantom{aa} \\
  Polytechnique Montréal, Canada \\
  \texttt{theo.guyard@polymtl.ca} \\
  \And
  Thibaut Vidal \\
  SCALE-AI Chair and CIRRELT \\
  Polytechnique Montréal, Canada \\
  \texttt{thibaut.vidal@polymtl.ca} \\
}

\begin{document}

\maketitle

\begin{abstract}
  Ensemble models achieve state-of-the-art performance on prediction tasks, but usually require aggregating a large number of weak learners. This can hinder deployment, interpretability, and downstream tasks such as robustness verification. Remedies to this issue fall into two main camps: pruning, which discards redundant learners, and compression, which generates new ones from scratch. We introduce \textsc{Pace}, a framework that interleaves these paradigms in a two-phase strategy. First, new learners are actively generated via a theoretically grounded procedure to enhance the diversity of the initial ensemble. When no more relevant learners can be found, a second phase of pruning is performed on this enriched ensemble. During both operations, \textsc{Pace} allows fine control on the faithfulness to the original ensemble. Experiments show that our method outperforms prior pruning and compression methods while offering principled control of faithfulness guarantees.
\end{abstract}

\input{sections/introduction}
\input{sections/framework}
\input{sections/outerloop}
\input{sections/innerloop}
\input{sections/experiments}
\input{sections/conclusions}

\bibliographystyle{plainnat}
\bibliography{references}

\clearpage

\appendix
\crefalias{section}{appendix}
\input{sections/supp-literature}
\input{sections/supp-cg}

\input{sections/supp-oracle}
\input{sections/supp-datasets}

\input{sections/supp-results}

\end{document}

%% file: headings/packages.tex
\usepackage{xparse}

\usepackage{amsmath}
\usepackage{amsfonts}
\usepackage{amsopn}
\usepackage{amssymb}
\usepackage{amsthm}
\usepackage{bbm}

\usepackage[utf8]{inputenc}
\usepackage[T1]{fontenc}
\usepackage[hidelinks]{hyperref}
\usepackage{url}
\usepackage{microtype}
\usepackage{soul}
\usepackage{lipsum}
\usepackage{cleveref}
\usepackage{rotating}
\usepackage{multirow}
\usepackage{siunitx}
\usepackage{enumitem}
\usepackage{todonotes}
\usepackage[acronym]{glossaries}
\usepackage{tcolorbox}
\usepackage{dirtytalk}
\usepackage{placeins}

\usepackage{booktabs}
\usepackage{tikz}
\usetikzlibrary{calc,shapes,arrows,positioning}
\usepackage{varwidth}
\usepackage{subcaption}
\usepackage{pgfplots}
\usepackage{pgfplotstable}
\usepgfplotslibrary{groupplots,fillbetween}
\usepgfplotslibrary{statistics}
\pgfplotsset{compat=1.17}

%% file: headings/macros.tex
\newcommand{\majorityvote}{V}

\newcommand{\initpredict}{y^o}
\newcommand{\initlearners}{\boldsymbol{h}^o}
\newcommand{\initlearnerweights}{\boldsymbol{w}^o}
\newcommand{\initlearnerweight}[1]{w^o_{#1}}
\newcommand{\initsamplespace}{\mathcal{X}^o}

\newcommand{\learnerset}{\mathcal{H}}
\newcommand{\learner}[1]{h_{#1}}
\newcommand{\learnerweight}[1]{w_{#1}}

\newcommand{\learners}{\boldsymbol{h}}
\newcommand{\learnerweights}{\boldsymbol{w}}

\newcommand{\samplespace}{\mathcal{X}}
\newcommand{\sampledim}{n}

\newcommand{\sample}[1]{\mathbf{x}_{#1}}

\newcommand{\sampleweight}[1]{v_{#1}}

\newcommand{\predictset}{\mathcal{Y}}
\newcommand{\predict}[1]{y_{#1}}

\newcommand{\conflevel}{\eta}
\newcommand{\plauslevel}{\delta}

\newcommand{\R}{\mathbb{R}}
\newcommand{\N}{\mathbb{N}}

\newcommand{\0}{\mathbf{0}}
\newcommand{\1}{\mathbf{1}}

\newcommand{\card}[1]{|#1|}
\DeclareMathOperator*{\argmax}{arg\,max}

\newtheorem{proposition}{Proposition}

\Crefname{lemma}{Lemma}{Lemmas}
\Crefname{proposition}{Proposition}{Propositions}
\Crefname{remark}{Remark}{Remarks}

\if@final
  \newcommand{\AddTodo}[1]{}

  \newcommand{\ProvideEditionMacros}[2]{%
    \expandafter\NewDocumentCommand\csname Add#1\endcsname{gg}{}%
    \expandafter\NewDocumentCommand\csname Rem#1\endcsname{gg}{}%
    \expandafter\NewDocumentCommand\csname Sup#1\endcsname{gg}{}%
  }
\else
  \newcommand{\AddTodo}[1]{\textcolor{red}{[Todo: #1]}}

  \newcommand{\ProvideEditionMacros}[2]{%
    \expandafter\NewDocumentCommand\csname Add#1\endcsname{gg}{\textcolor{#2}{##1}}%
    \expandafter\NewDocumentCommand\csname Rem#1\endcsname{gg}{\textcolor{#2}{[#1: ##1]}}%
    \expandafter\NewDocumentCommand\csname Sup#1\endcsname{gg}{\textcolor{#2}{\st{##1}}}%
  }
\fi

%% file: sections/introduction.tex
\section{Introduction}

Ensemble modeling is a widely used machine learning technique where several weak learners are aggregated into a single strong predictor \cite{sagi2018ensemble}. It has achieved state-of-the-art performance across many applications, especially on tabular data~\citep{MAKRIDAKIS2022}, capturing complex relationships while remaining robust to overfitting. Formally, an ensemble consists of a collection~$\learners \subseteq \learnerset$ of learners from a prescribed family, such as decision trees. Each learner $\learner{} \in \learnerset$ encodes functions $\learner{\predict{}} : \samplespace \to \R_+$ assigning a score to possible prediction labels $\predict{} \in \predictset$ over a sample space~$\samplespace \subseteq \R^{\sampledim}$, larger values indicating stronger support. The ensemble assigns a prediction label to any sample $\sample{} \in \samplespace$ queried via a majority vote parameterized by weights $\learnerweights \subseteq \R_+$ as
\begin{equation}
    \label{eq:majority_vote}
    \textstyle
    \majorityvote_{\learners}(\learnerweights,\sample{}) \in \argmax_{\predict{} \in \predictset}
    \sum_{\learner{} \in \learners} \learnerweight{\learner{}} \cdot \learner{\predict{}}(\sample{}).
\end{equation}
Recent machine learning competitions have shown that strong predictive performance with ensemble models often requires a large number of learners~\citep{MAKRIDAKIS2022}. At the same time, popular training methods such as random forests and boosting often produce redundant learners due to their heuristic nature~\cite{friedman2000additive}. This redundancy increases memory footprint and inference time, which are critical limitations in latency-sensitive and resource-constrained deployment settings~\citep{10153433,10.1145/3581759,pmlr-v70-kumar17a}. These issues are amplified at scale, where even small per-query overheads can translate into substantial computational and energy costs. Large ensembles also complicate downstream audit and explanation tasks, including the computation of optimal counterfactual explanations~\citep{pmlr-v139-parmentier21a} and robustness verification~\citep{NEURIPS2019_cd9508fd}. They may further reduce interpretability by increasing decision complexity, measured as the number of conditions evaluated to produce a prediction~\citep{10.1145/3600211.3604664}. Finally, they can leak more information to adversaries: for example, trained random forests can be used to reconstruct much of their training data, with attack success strongly depending on the number and depth of trees~\citep{pmlr-v235-ferry24a}.

\paragraph{Ensemble Pruning and Compression.}

To address these issues, several studies have proposed \emph{pruning} strategies that assign a new set of sparse weights to a trained ensemble model, aiming to discard redundant learners \citep{zhang2006ensemble}. A central requirement in this process is to preserve the predictive behavior of the original ensemble, either empirically on a finite set of samples, or formally over all samples in a region of interest. We refer to the latter requirement as \emph{faithfulness}, and focus on methods that provide such guarantees. Although recent pruning approaches can certify faithfulness~\citep{emine2025free}, their pruning potential is limited by the fact that they can only reweight or discard existing learners. Another line of research instead focuses on \emph{compressing} the initial ensemble by the active generation of new learners. Some approaches in this direction can also preserve faithfulness, but typically require producing learners that are more complex than the original ones and no longer belong to the prescribed family~$\learnerset$~\citep{vidal2020born}. Other approaches allow sticking to the original family of learners \citep{zhou2016interpreting}, but achieve compression without faithfulness guarantees, thereby potentially altering prediction and generalization capabilities. A thorough review of these techniques is provided in Appendix~\ref{sec:related_works}.

\paragraph{Contributions.}

This paper introduces \textsc{Pace} (Prune-And-Compress Ensemble), a novel framework that unifies pruning and compression techniques for ensemble models while providing flexible, yet rigorous, control over faithfulness enforcement. Our method is built upon the following contributions:
\begin{enumerate}
    \item We show that pruning and compression are complementary rather than competing paradigms. Based on this observation, we propose a unified framework that first actively generates new learners to help compress the ensemble, and then prunes the redundant ones generated along the way.
    \item We leverage column generation to provide a principled strategy for generating new learners along with a stopping criterion. Thanks to this design, the enriched ensemble has greater potential for subsequent pruning.
    \item Within this unified framework, we extend prior notions of faithfulness to target only regions that are meaningful in practice. This enables a trade-off between ad hoc approximations of the initial decision boundary with no faithfulness guarantees, and exact imitation, which can otherwise limit compression.
    \item Numerical experiments suggest that \textsc{Pace} achieves stronger compression and pruning performance than prior methods, while providing finer control over faithfulness.
\end{enumerate}
Our exposition is organized as follows. \Cref{sec:framework} gives a global overview of \textsc{Pace}, with the core methodologies and definitions. Technical details are presented in \Cref{sec:outer,sec:inner}. The performance of our method on practical problems is then assessed in \Cref{sec:experiments}. Finally, \Cref{sec:conclusions} discusses limitations and future research directions. Appendices gather additional technical details, material to reproduce our experiments, and further numerical analyzes. 

%% file: sections/framework.tex
\section{Prune-and-Compress Framework for Ensemble Models}
\label{sec:framework}

The gold standard for obtaining a compact and faithful representation of a weighted ensemble~$(\initlearners,\initlearnerweights)$ is to identify a new pair $(\learners,\learnerweights)$ achieving the optimal value in the following problem:
\begin{equation}
    \label{prob:l0}
    \tag{$P$}
    \textstyle
    \min_{\learnerweights \geq \0} \ \|\learnerweights\|_0
    \ \ \text{s.t.} \ \ 
    \majorityvote_{\learners}(\learnerweights,\sample{}) = \majorityvote_{\initlearners}(\initlearnerweights,\sample{}) \ \ \forall \sample{} \in \initsamplespace,
\end{equation}
where the $\ell_0$-norm counts the number of nonzero weights, and the constraints enforce faithfulness over a region $\initsamplespace \subseteq \samplespace$ of interest. This ideal approach raises two key difficulties. First, it requires searching over possible subsets of learners in $\learnerset$, which itself can be very large or even infinite. Second, even for a fixed subset of learners, Problem~\eqref{prob:l0} remains challenging because of the combinatorial nature of the $\ell_0$-norm objective and the large, potentially infinite, set of faithfulness constraints.

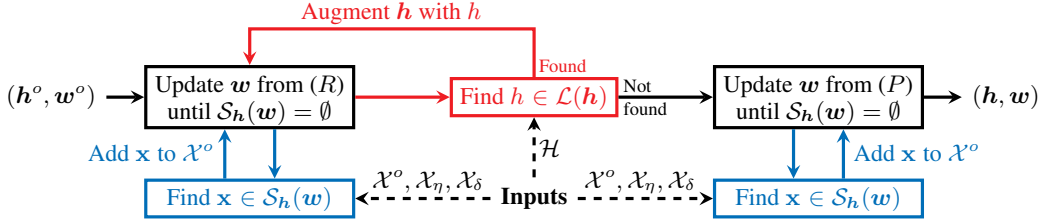
\begin{figure}[!htbp]
    \centering
    \input{figures/backbone}
    \caption{High-level description of \textsc{Pace} highlighting the active generation of improving learners (red) and the enforcement of faithfulness via separating sample generation (blue). The family of learners and the regions where faithfulness is enforced are tunable inputs of the method.}
    \label{fig:backbone}
\end{figure}

Pruning and compression techniques can both be viewed as surrogate strategies for approximately addressing this ideal task. On the one hand, pruning restricts its focus to the original ensemble $\learners=\initlearners$ and computes a one-shot solution of Problem \eqref{prob:l0} to obtain a sparser set of weights. However, fixing the ensemble inherently limits pruning capabilities. On the other hand, compression actively generates a new ensemble $\learners$, but disregards sparsification of the associated weights.
Both methods usually enforce faithfulness over the whole region $\samplespace=\initsamplespace$, which can lead to stringent constraints that degrade pruning and compression performances. Motivated by these observations, \textsc{Pace} combines the strengths of pruning and compression techniques, together with a finer control of faithfulness. Its main ingredients are outlined below and in \Cref{fig:backbone}, with further details given in \Cref{sec:outer,sec:inner}.

\paragraph{Method backbone.}
\textsc{Pace} is structured in two phases. First, it actively generates new learners to enrich the original ensemble, aiming to improve the ensemble's potential for downstream pruning task in a second phase, where a one-shot solution to Problem \eqref{prob:l0} is computed using the augmented ensemble. Specifically, \textsc{Pace} starts with $\learners=\initlearners$ and searches for any \emph{improving learner} in the set
\begin{equation}
    \label{eq:improving-learner-set}
    \mathcal{L}(\learners) = \{\learner{} \in \learnerset : \ell(\learners \cup \{\learner{}\}) < \ell(\learners)\},
\end{equation}
where a larger value of the measure $\smash{\ell : 2^{\learnerset} \to \mathbb{R}_+}$ indicates stronger potential for the downstream pruning operation. Instead of directly relying on the optimal value of the pruning Problem \eqref{prob:l0} to quantify $\ell(\learners)$, this measure is rather defined as the optimal value to its relaxation
\begin{equation}
    \label{prob:l1}
    \tag{$R$}
    \textstyle
    \min_{\learnerweights \geq \0} \ \|\learnerweights\|_1
    \ \ \text{s.t.} \ \ 
    \majorityvote_{\learners}(\learnerweights,\sample{}) = \majorityvote_{\initlearners}(\initlearnerweights,\sample{}) \ \ \forall \sample{} \in \initsamplespace,
\end{equation}
involving an $\ell_1$-norm surrogate. Stated otherwise, any $\learner{} \in \mathcal{L}(\learners)$ can be added to the current ensemble $\learners$ to improve the optimal value achieved by Problem~\eqref{prob:l1}. The main advantage of this definition is that this relaxation can be cast as a linear program. Building on column generation techniques~\citep{desaulniers2006column}, we provide a principled and efficient strategy to retrieve improving learners, together with a theoretically-grounded criterion to verify that $\mathcal{L}(\learners)=\emptyset$, leading to the following guarantee.
\begin{center}
    \begin{tcolorbox}[colframe=black,colback=white,sharp corners=all, boxsep=-0.5mm, width=0.9\linewidth, boxrule=1pt]
        \textbf{Pruning guarantee:} When $\mathcal{L}(\learners) = \emptyset$, no new learner in $\learnerset$ can be added to the current ensemble $\learners$ to improve the optimal value achieved by Problem \eqref{prob:l1}.
    \end{tcolorbox}
\end{center}
Once this condition is met, the enriched ensemble $\learners$ is guaranteed to contain all necessary learners to achieve the best possible optimal value in~\eqref{prob:l1}, and is thus expected to offer strong pruning capabilities via Problem \eqref{prob:l0}. \textsc{Pace} then carries out a pruning phase through a one-shot solution of Problem \eqref{prob:l0} to assign the sparsest possible set of weights in this enriched ensemble. Overall, \textsc{Pace} benefits from more flexibility in the choice of learners than pruning techniques, and also accounts for sparsification of their weights, which is not considered in compression techniques.

\paragraph{Faithfulness management.}
Instead of enforcing faithfulness over an empirical set of samples $\initsamplespace \subseteq \samplespace$ when solving Problems \eqref{prob:l0} and \eqref{prob:l1}, \textsc{Pace} enforces more global faithfulness guarantees. After solving the problem at hand, the method searches for \emph{separating samples} in the region
\begin{equation}
    \label{eq:separating-samples}
    \mathcal{S}_{\learners}(\learnerweights) = \{\sample{} \in \samplespace_{\conflevel} \cap \samplespace_{\plauslevel} :
    \majorityvote_{\learners}(\learnerweights,\sample{}) \neq \majorityvote_{\initlearners}(\initlearnerweights,\sample{})\},
\end{equation}
where the predictions of the current ensemble $(\learners,\learnerweights)$ and the initial one $(\initlearners,\initlearnerweights)$ disagree. Any separating sample found is added to $\initsamplespace$, and the problem is solved again to update the weights in the current pair $(\learners,\learnerweights)$. This process is repeated until no more separating sample can be found. Unlike prior approaches \citep{emine2025free} retrieving separating samples over the entire space $\samplespace$, \textsc{Pace} controls faithfulness guarantees over a region of practical interest, defined from the intersection of the region
\begin{equation}
    \textstyle
    \samplespace_{\conflevel}=\big\{
        \sample{} \in \samplespace
        \mid
        \textstyle
    \sum_{\learner{} \in \initlearners} \ 
    \initlearnerweight{\learner{}} \cdot (\learner{\initpredict}(\sample{}) -\learner{\tilde{\predict{}}}(\sample{})) > \conflevel \ \ \forall \tilde{\predict{}} \neq \initpredict
    \big\}
\end{equation}
capturing samples where the initial ensemble takes a prediction $\initpredict=\majorityvote_{\initlearners}(\initlearnerweights,\sample{})$ with a \emph{confidence} margin $\conflevel \geq 0$ compared to alternative classes, and the region
\begin{equation}
    \textstyle
    \samplespace_{\plauslevel}=\big\{
        \sample{} \in \samplespace
        \mid
        \sum_{\learner{} \in \mathcal{T}} b_{\learner{}}(\sample{}) \geq \plauslevel
    \big\}
\end{equation}
where $b_{\learner{}}(\sample{}) \geq 0$ denotes \emph{plausibility} score of a sample $\sample{} \in \samplespace$ with respect to the data-distribution. This score is computed through trees $\learner{} \in \mathcal{T}$ of an isolation forest that is broadly employed in the literature to identify out-of-distribution samples~\citep{liu2008isolation,liu2012isolation}. Intuitively, outliers are typically isolated within sparse leaves and after only a few splits in a decision tree, therefore receiving low plausibility scores.
This faithfulness control is motivated by two complementary considerations. On the one hand, models with comparable accuracy may still disagree on individual predictions, so exact faithfulness of the original ensemble in low-confidence regions can preserve an arbitrary choice among similarly valid alternatives. On the other hand, plausibility restriction prevents faithfulness constraints from being driven by out-of-distribution samples, where preserving the original ensemble's predictions is of limited practical relevance. In this way, relaxing faithfulness over irrelevant regions allows for potentially stronger pruning capabilities during \textsc{Pace}, with the following guarantee:
\begin{center}
    \begin{tcolorbox}[colframe=black,colback=white,sharp corners=all, boxsep=-0.5mm, width=0.9\linewidth, boxrule=1pt]
    \textbf{Faithfulness guarantee:} When $\mathcal{S}_{\learners}(\learnerweights) = \emptyset$, the solution $(\learners,\learnerweights)$ of the Problem \eqref{prob:l0} or \eqref{prob:l1} at hand is faithful to the initial ensemble $(\initlearners,\initlearnerweights)$ over the region $\samplespace_{\conflevel} \cap \samplespace_{\plauslevel}$.
    \end{tcolorbox}
\end{center}
This novel definition provides finer control over the region where faithfulness is enforced, potentially achieving stronger pruning capabilities during \textsc{Pace}, based on two meaningful parameters, $\conflevel \geq 0$ and $\plauslevel \geq 0$. Setting $\conflevel=\plauslevel=0$ recovers the case where faithfulness is enforced over the whole space~$\samplespace$. In practice, \textsc{Pace} opts for a constraint programming formulation of the region $\mathcal{S}_{\learners}(\learnerweights)$, allowing to retrieve separating samples or prove that no more exist with enhanced computational performance compared state-of-the-art formulations previously proposed in the literature for similar tasks.

%% file: figures/backbone.tex
\tikzstyle{block} = [draw, rectangle, fill=white, minimum width=2.75cm, align=center]
\tikzstyle{virtual} = [coordinate]

\begin{tikzpicture}[font=\small, very thick, >=stealth, auto, node distance=1.25cm]

    \node (initial) {$(\initlearners,\initlearnerweights)$};
    \node [block, right=0.5cm of initial] (l1) {Update $\learnerweights$ from \eqref{prob:l1} \\ until $\smash{\mathcal{S}_{\learners}(\learnerweights) = \emptyset}$};
    \node [block, minimum width=2cm, Red, fill=white, right=of l1] (trainer) {Find $\learner{} \in \smash{\mathcal{L}(\learners)}$};
    \node [block, right=of trainer] (l0) {Update $\learnerweights$ from \eqref{prob:l0} \\ until $\smash{\mathcal{S}_{\learners}(\learnerweights) = \emptyset}$};
    \node [right=0.5cm of l0] (final) {$(\learners,\learnerweights)$};

    \draw [->] (initial) -- (l1);
    \draw [->, Red] (l1) -- (trainer);
    \draw [->] (trainer) -- (l0);
    \draw [->] (l0) -- (final);

    \node [block, NavyBlue, fill=white, below=0.65cm of l1] (l1oracle) {Find $\sample{} \in \smash{\mathcal{S}_{\learners}(\learnerweights)}$};
    \draw [->, NavyBlue] ($(l1.south)+(0.33cm,0)$) -- ($(l1oracle.north)+(0.33cm,0)$);
    \draw [->, NavyBlue] ($(l1oracle.north)-(0.33cm,0)$) -- ($(l1.south)-(0.33cm,0)$) node [midway, left] {Add $\sample{}$ to $\initsamplespace$};
   
    \node [block, NavyBlue, fill=white, below=0.65cm of l0] (l0oracle) {Find $\sample{} \in \smash{\mathcal{S}_{\learners}(\learnerweights)}$};
    \draw [->, NavyBlue] ($(l0.south)-(0.33cm,0)$) -- ($(l0oracle.north)-(0.33cm,0)$);
    \draw [->, NavyBlue] ($(l0oracle.north)+(0.33cm,0)$) -- ($(l0.south)+(0.33cm,0)$) node [midway, right] {Add $\sample{}$ to $\initsamplespace$};

    \node [Red, anchor=south west, font=\scriptsize, xshift=-2pt, yshift=-2pt] at (trainer.north) {Found};
    \node [anchor=south west, font=\scriptsize, xshift=-2pt, yshift=-2pt] at (trainer.east) {Not};
    \node [anchor=north west, font=\scriptsize, xshift=-2pt, yshift=2pt] at (trainer.east) {found};
    \node [virtual] (learner-loopback) at ($(trainer)!0.5!(l1)+(0,0.9cm)$) {};
    \draw [->, Red] (trainer) |- (learner-loopback) -| (l1);
    \node [anchor=south, Red, yshift=-2pt] at (learner-loopback) {Augment $\learners$ with $\learner{}$};

    \node[align=center] (data) at ($(l1oracle)!0.5!(l0oracle)$) {\textbf{Inputs}};
    \draw [->, dashed] (data) -- (l1oracle) node [midway, above,xshift=2pt,yshift=-2pt] {$\initsamplespace,\samplespace_{\conflevel},\samplespace_{\plauslevel}$};
    \draw [->, dashed] (data) -- (l0oracle) node [midway, above,xshift=-2pt,yshift=-2pt] {$\initsamplespace,\samplespace_{\conflevel},\samplespace_{\plauslevel}$};
    \draw [->, dashed] (data) -- (trainer) node [midway, right,xshift=-2pt] {$\learnerset$};
\end{tikzpicture}

%% file: sections/outerloop.tex
\section{Building Improving Learners via Column Generation}
\label{sec:outer}

A key feature of \textsc{Pace}, which distinguishes it from previous approaches, is that it first augments the initial ensemble~$\initlearners$ with improving learners before performing the final pruning step. The goal is to produce an enriched ensemble with learners that make stronger compression possible later on. We measure this potential through the optimal value of Problem \eqref{prob:l1}, and use column generation to guide the selection of new learners in a principled way.

\paragraph{Improving learners detection.}
From Definition \eqref{eq:improving-learner-set}, an improving learner is any new $\learner{} \in \learnerset$ that can be added to the current ensemble $\learners$ being augmented during the first phase of \textsc{Pace} to reduce the optimal value achieved by Problem \eqref{prob:l1}. To identify them, we observe that the latter problem can be seen as a restricted version of the extended formulation
\begin{equation}
    \label{prob:l1-ext}
    \tag{$R_{\text{ext}}$}
    \textstyle
    \min_{\learnerweights \geq \0} \ \|\learnerweights\|_1
    \ \ \text{s.t.} \ \ 
    \majorityvote_{\learnerset}(\learnerweights,\sample{}) = \majorityvote_{\initlearners}(\initlearnerweights,\sample{}) \ \ \forall \sample{} \in \initsamplespace,
\end{equation}
involving all possible learners from the family $\learnerset$.\footnote{Our method accommodates scenarios where $\learnerset$ is of infinite cardinality, as the theoretical results from the semi-infinite linear programming literature allow to adapt column generation to such infinite-dimensional settings \citep{lopez2007semi}.} By design, this extended formulation achieves a better optimal value than Problem \eqref{prob:l1} since it benefits from more flexibility in the choice of learners. Hence, generating an improving learner amounts to checking whether there exists some $\learner{} \in \learnerset$ such that the inequalities $\ell(\learnerset) \leq \ell(\learners \cup \learner{}) < \ell(\learners)$ hold, reminding that $\ell(\cdot)$ denotes the optimal value of Problem \eqref{prob:l1} for a given ensemble of learners. No more improving learners exist once $\ell(\learnerset) = \ell(\learners)$. Interestingly, Problem \eqref{prob:l1} can be cast into a linear program as
\begin{equation}
    \label{prob:l1-lp}
    \tag{$R$}
    \left\{
    \begin{array}{lll}
        \min_{\learnerweights \geq \0} & \sum_{\learner{} \in \learners} \learnerweight{\learner{}} \\
        \text{s.t.} & \sum_{\learner{} \in \learners}
        \learnerweight{\learner{}}
        \big(
        \learner{\initpredict}(\sample{})
        -
        \learner{\predict{}}(\sample{})
        \big) \geq 1 &\forall \sample{} \in \initsamplespace, \ \forall \predict{} \neq \initpredict,
    \end{array}
    \right.
\end{equation}
where $\initpredict = \majorityvote_{\initlearners}(\initlearnerweights,\sample{})$ denotes the class predicted by the original ensemble. Each constraint enforces faithfulness for a sample $\sample{} \in \initsamplespace$ through the expanded expression of the majority vote $\majorityvote_{\learners}(\learnerweights,\sample{})=\majorityvote_{\initlearners}(\initlearnerweights,\sample{})$ in~\eqref{eq:majority_vote}. Note that the right-hand side in the constraints can be set arbitrarily to one since the problem is scale-invariant with respect to the variable $\learnerweights \geq \0$. A similar linear program reformulation holds for the extended formulation~\eqref{prob:l1-ext}. Building on column generation theory, we then establish the following result.

\begin{proposition}
    \label{prop:pricing}
    Consider Problem~\eqref{prob:l1-lp} constructed for some ensemble $\learners \subseteq \learnerset$, and define
    \begin{equation}
        \label{prob:pricing}
        c^\star = \min_{\learner{} \in \learnerset} \ 1 - \sum_{\sample{} \in \initsamplespace}
        \sum_{\predict{} \neq \initpredict}
        \sampleweight{\sample{},\predict{}} \cdot 
        \big(
            \learner{\initpredict}(\sample{}) -
            \learner{\predict{}}(\sample{})
        \big),
    \end{equation}
    where $\sampleweight{\sample{},\predict{}} \geq 0$ is the optimal solution in the dual of Problem \eqref{prob:l1-lp} associated with the constraint defined for some pair $(\sample{},\predict{}) \in \initsamplespace \times \predictset$. If $c^{\star} < 0$, any minimizer to Problem \eqref{prob:pricing} is an improving learner verifying $\learner{} \in \mathcal{L}(\learners)$. Otherwise, we have $\mathcal{L}(\learners)=\emptyset$.
\end{proposition}

Proof of this property is provided in Appendix \ref{app:proof-pricing}, where we show that the optimal value of the \emph{pricing} Problem \eqref{prob:pricing} determines the marginal modification in the measure $\ell(\cdot)$. If this quantity is strictly negative, any solution $\learner{} \in \learnerset$ ensures a strict decrease $\ell(\learners \cup \learner{}) < \ell(\learners)$, whereas when it is non-negative, no new improving learner can be generated and we have $\ell(\learnerset) = \ell(\learners)$, i.e., the optimal value of Problem \eqref{prob:l1-lp} coincides with that of the extended formulation \eqref{prob:l1-ext}. Solving this pricing problem is then sufficient to identify improving learners during \textsc{Pace} or to detect that no more remain, meaning that the pruning guarantee stated in \Cref{sec:framework} is met.

\paragraph{Pricing problem as a learning task.}
At first glance, the pricing Problem~\eqref{prob:pricing} may appear challenging, as it involves optimizing over all possible learners in the family $\learnerset$. However, borrowing a shift in perspective from the LP-boosting literature \citep{Demiriz2002}, we observe that it can be reformulated as
\begin{equation}
    \label{prob:pricing-reformulated}
    \max_{\learner{} \in \learnerset} \ \sum_{\sample{} \in \initsamplespace}
        \sum_{\predict{} \neq \initpredict}
        \sampleweight{\sample{},\predict{}} \cdot 
        \big(
            \learner{\initpredict}(\sample{}) -
            \learner{\predict{}}(\sample{})
        \big),
\end{equation}
by switching the objective direction and discarding the constant term. This reformulation can be interpreted as a search for a new learner $\learner{} \in \learnerset$ that maximizes the margin between the score assigned to the class $\initpredict = \majorityvote_{\initlearners}(\initlearnerweights,\sample{})$ predicted by the original ensemble and that assigned to any alternative class $\predict{} \neq \initpredict$ across all samples in $\initsamplespace$, weighted by the value of solutions in the dual of Problem~\eqref{prob:l1-lp}. Consequently, evaluating the condition in \Cref{prop:pricing} can be equivalently performed by training a new learner $\learner{} \in \learnerset$ on a re-weighted version of $\initsamplespace$ with a particular loss function. This interpretation is particularly clear for binary classification where~$\predictset=\{-1,+1\}$. In this case, the pricing problem can be equivalently expressed as
\begin{equation}
    \label{prob:pricing-binary}
    \max_{\learner{} \in \learnerset}
    \sum_{\sample{} \in \initsamplespace}
    \sampleweight{\sample{},-\initpredict} \cdot \learner{\initpredict}(\sample{}),
\end{equation}
and corresponds to a classical weighted prediction score maximization for a target label \citep[Sec.~4]{Demiriz2002}.

The pricing Problem \eqref{prob:pricing} must be solved optimally to guarantee that the condition $\mathcal{L}(\learners) = \emptyset$ terminating the generation of improving learners is correctly evaluated via \Cref{prop:pricing} during \textsc{Pace}. For some families of learners, exact algorithmic methods \citep{demirovic2023blossom} exist, but they might be computationally intensive. Alternatively, it is possible to replace the exact pricing with a heuristic, such as CART \citep{breiman2017classification}, yielding an approximate solution to the pricing Problem \eqref{prob:pricing-reformulated} at a low computational cost. Note that the heuristic or exact nature of the solution to this pricing problem solely drives the enrichment of the ensemble performed during the first phase of \textsc{Pace}, but does not impact its faithfulness guarantees.

%% file: sections/innerloop.tex
\section{Retrieving Separating Samples to Achieve Faithfulness}
\label{sec:inner}

Iteratively, after each solution of Problem \eqref{prob:l0} or \eqref{prob:l1},
\textsc{Pace} searches for separating samples using constraint programming (CP) techniques. For each pair $(\initpredict,\predict{})\in\predictset^2$ of distinct labels, it searches for any $\sample{} \in \mathcal{S}_{\learners}(\learnerweights)$ such that $\majorityvote_{\initlearners}(\initlearnerweights,\sample{}) = \initpredict$ and $\majorityvote_{\learners}(\learnerweights,\sample{})=\predict{}$. It either returns one such separating sample per pair, or certifies that none exists, i.e., $\mathcal{S}_{\learners}(\learnerweights)=\emptyset$ and the algorithmic cycle stops. More precisely, this CP-based separation approach builds on the mixed-integer formulation introduced by ~\citet{emine2025free} for a similar task for families $\learnerset$ of decision tree learners. However, it differs in two important ways: (i) it uses a constraint-based separation enforcement and avoids cumbersome linearization of logical constraints to get the most out of CP primitives, accelerating separating sample retrieval by several orders of magnitude, and (ii) it accommodates our novel notion of faithfulness that incorporates the notion of confidence of the original ensemble and the plausibility of the separating samples. We present the CP formulation below and refer to Appendix~\ref{appendix:oracle} for further technical details.

\textbf{Modeling ensemble predictions.}
A key ingredient in our CP-based formulation of the region $\mathcal{S}_{\learners}(\learnerweights)$ is a new mathematical model for the prediction score $\learner{\predict{}}:\samplespace\to\predictset$ of learners in the family $\learnerset$ of decision trees. For such learners, denote by $\mathcal{V}_{\learner{}}$ the set of leaves of a decision tree $\learner{} \in \learnerset$, by $\smash{\Gamma_{\learner{}v}^+}$ (resp. $\smash{\Gamma_{\learner{}v}^-}$) the set of index-scalar pairs $(i,a)$ such that the $i$-th entry of some $\sample{} \in \samplespace$ must be greater (resp. smaller) than $a \in \R$ to reach leaf $v \in \mathcal{V}_{\learner{}}$, and by $v_{\predict{}}\in\R_+$ the leaf prediction score for each label $\predict{} \in \predictset$.
Then, a relation $r_{\learner{}\predict{}}=\learner{\predict{}}(\sample{})$ can be modeled for any sample $\sample{} \in \samplespace$ as
\begin{subequations}
    \label{oracle:base}
    \begin{align}
        \label{oracleconstr:flow_predict}
        &&
        \textstyle
        \sum_{v \in \mathcal{V}_{\learner{}}}v_{\predict{}}  z_{\learner{}v}&=r_{\learner{}\predict{}} \\
        &&
        \label{oracleconstr:flow_uniqueness}
        \textstyle
        \sum_{v \in \mathcal{V}_{\learner{}}} z_{\learner{}v}&=1 \\
        &&
        \label{oracleconstr:flow_vars_consistency}
        \textstyle
        z_{\learner{}v}&=1\textstyle\implies \big(\bigwedge_{(i,a)\in\Gamma_{\learner{}v}^{+}} x_i>a\big)\land\big(\bigwedge_{(i,a)\in\Gamma_{\learner{}v}^{-}} x_i\leq a\big) && \forall v \in \mathcal{V}_{\learner{}}
    \end{align}
\end{subequations}
using a set of binary variables $z_{\learner{}v} \in \{0,1\}$ indicating whether the sample falls into leaf $v \in \mathcal{V}_{\learner{}}$ of the tree $\learner{} \in \learnerset$. Constraints~\eqref{oracleconstr:flow_uniqueness}--\eqref{oracleconstr:flow_vars_consistency} ensure that a unique leaf is reached and that branching conditions are consistent, while constraint \eqref{oracleconstr:flow_predict} models the tree prediction score. This formulation is particularly suited for CP solvers since they natively support the logical implications in \eqref{oracleconstr:flow_vars_consistency} through efficient reification constraints \citep{jefferson2010implementing}. Moreover, expression \eqref{oracleconstr:flow_uniqueness} can be cast as a global \texttt{ExactlyOne} constraint, often largely enhancing their efficiency~\citep{rossi2006handbook}. In our CP-based formulation of the separation problem, we integrate one such set of constraints \eqref{oracleconstr:flow_predict} for each initial and new learner in $\learner{} \in \initlearners \cup \learners$.

\textbf{Constraint-based separation.} 
Building on the mathematical model \eqref{oracle:base} for the prediction score of tree-based learners, we propose to enforce separation via the constraints
\begin{align}
    \label{oracleconstr:initial_class}
    \textstyle
    \sum_{\learner{} \in \initlearners} \ 
    \initlearnerweight{\learner{}} \cdot (r_{\learner{}\initpredict} -r_{\learner{}\tilde{\predict{}}}) &> 0 \quad \forall \tilde{\predict{}} \neq \initpredict \\
    \label{oracleconstr:separation_constraint}
    \textstyle\sum_{\learner{} \in \learners} \ 
    \learnerweight{\learner{}} \cdot (r_{\learner{}\predict{}} -r_{\learner{}\initpredict}) &> 0
\end{align}
defined for a pair $(\initpredict,\predict{}) \in \predictset^2$ of distinct prediction labels. While constraint \eqref{oracleconstr:initial_class} enforces the prediction relation $\initpredict = \majorityvote_{\initlearners}(\initlearnerweights,\sample{})$ for the initial ensemble, constraint \eqref{oracleconstr:separation_constraint} promotes an alternate class $\predict{} = \majorityvote_{\learners}(\learnerweights,\sample{})$ for the current ensemble processed by \textsc{Pace}. Hence, any $\sample{} \in \samplespace$ feasible for this problem is a separating sample. Prior state-of-the-art formulations of separation \citep{emine2025free} rather opted for maximizing the left-hand side of constraint~\eqref{oracleconstr:separation_constraint}, thereby seeking a maximal disagreement in predictions. This usually increases runtime because both separation and faithfulness require solving the optimization problem to optimality: first to find the maximal disagreement sample, and then to prove that no sample with positive objective exists. In contrast, we opt for a constraint-based separation returning \emph{any} separating sample. This paradigm reduces the search of separation samples to a feasibility problem: any feasible solution is a separating sample, and infeasibility certifies faithfulness for the considered pair of distinct labels.

\textbf{Confidence and plausibility regions.}
With constraints \eqref{oracleconstr:initial_class}-\eqref{oracleconstr:separation_constraint} only, separating samples can be recovered from the entire space $\samplespace$. To further control their search to the confidence region $\samplespace_{\conflevel}$ involved in the definition of $\mathcal{S}_{\learners}(\learnerweights)$, a simple modification of the right-hand-side of constraint \eqref{oracleconstr:initial_class} as 
\begin{align}
    \textstyle
    \sum_{\learner{} \in \initlearners} \ 
    \initlearnerweight{\learner{}} \cdot (r_{\learner{}\initpredict} -r_{\learner{}\tilde{\predict{}}}) &> \conflevel \quad \forall \tilde{\predict{}} \neq \initpredict
\end{align}
is required. Hence, the notion of confidence of the initial ensemble readily integrates our initial CP-based formulation, without any new variables and constraints. Besides, the search of separating samples can be further controlled to the plausibility region $\samplespace_{\plauslevel}$ involved in the definition of $\mathcal{S}_{\learners}(\learnerweights)$ by introducing analogous constraints to \eqref{oracle:base} for each tree of a given isolation forest $\mathcal{T}$ capturing the plausibility scores, together with the additional constraint
\begin{align}
    \label{eq:plausibility_score}
    \textstyle
    \sum_{\learner{} \in \mathcal{T}}
    \sum_{v \in \mathcal{V}_{\learner{}}}
    b_{\learner{}v} \cdot z_{\learner{}v} \geq \delta,
\end{align}
where the coefficients $b_{\learner{}v}\geq 0$ encode the corrected path length associated with each leaf $v \in \mathcal{V}_{\learner{}}$ of each tree $\learner{} \in \mathcal{T}$ in the isolation forest. They can be pre-computed since $\mathcal{T}$ is fixed throughout \textsc{Pace}.

%% file: sections/experiments.tex
\section{Numerical Experiments}
\label{sec:experiments}

We now empirically evaluate \textsc{Pace} and demonstrate that: (i) our constraint programming-based separation problem formulation of faithfulness outperforms current objective-based ones, (ii) the active generation of learners enables stronger capabilities compared to pruning-only methods, and (iii) controlling faithfulness over a region of practical interest improves pruning performance compared to strict faithfulness enforcement. Our experimental setup and the datasets used are described in \Cref{app:datasets}. We study random forest ensembles constructed from tree learners. Complementary results with AdaBoost ensembles and further numerical analyses are provided in \Cref{app:comlementary_results}.

\begin{figure}[hbtp]
    \centering

    \begin{subfigure}{\textwidth}
        \centering
        \includegraphics[clip, trim=0.1cm 0.04cm 0.1cm 0.1cm,
                         width=0.65\textwidth]{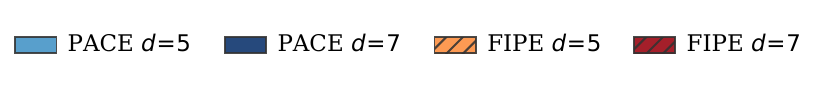}
    \end{subfigure}
    \vspace{-2.4em}

    \begin{subfigure}[t]{0.7\textwidth}
        \centering
        \includegraphics[width=\textwidth]{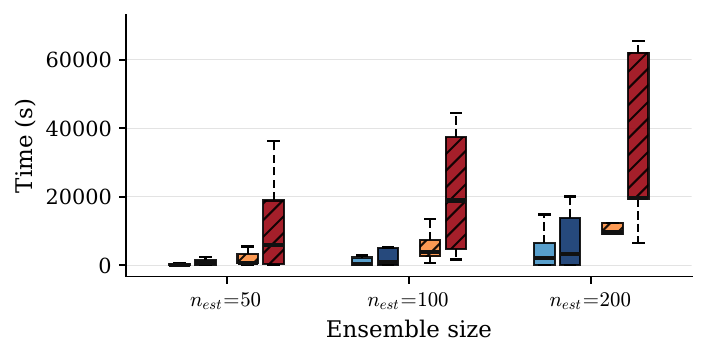}
    \end{subfigure}\hfill
    \caption{Time to achieve global faithfulness for Problem~\eqref{prob:l1} via the constraint-based formulation used in \textsc{Pace} and the objective-based one used in \textsc{Fipe} for retrieving separating samples. Boxes show random forest ensembles of size $n_{\text{est}} \in \{50, 100, 200\}$, with color shade encoding tree depth $d \in \{5, 7\}$. For a fair comparison, results are aggregated on the datasets listed in \Cref{app:datasets} (11 in total) where both \textsc{Fipe} and \textsc{Pace} terminate within the 20-hour time limit, i.e. on 9 data sets for $n_{\text{est}}=50$, 8 datasets for $n_{\text{est}}=100$, and 5 datasets for $n_{\text{est}}=200$. In practice, \textsc{Pace} terminated over a larger number of datasets, namely {11} for $n_{\text{est}}=50$, {11} for $n_{\text{est}}=100$, and {10} for $n_{\text{est}}=200$.}
    \label{fig:compu_time_rf}
\end{figure}

\paragraph{Result 1: The constraint-based separation formulation underlying \textsc{Pace} achieves significant speedups over the state of the art.}
We first study how the constraint-based separation problem used in \textsc{Pace} compares to the state-of-the-art objective-based formulation of \citet{emine2025free} in their \textsc{Fipe} method to retrieve separating samples over the entire feature space $\samplespace$. To this end, we iteratively solve Problem \eqref{prob:l1} and generate separating samples until faithfulness is guaranteed globally over $\samplespace = \R^{\sampledim}$, setting $\learners=\initlearners$ and initializing $\initsamplespace$ with the training samples of the initial ensemble~$(\initlearners, \initlearnerweights)$. The same solver is used for each solution of Problem \eqref{prob:l1}, so as to only reflect the performance of the separating sample generation. \Cref{fig:compu_time_rf} shows that our constraint-based formulation yields substantial computational gains over the objective-based approach of \citet{emine2025free}. In the worst case with random forests ($n_{\text{est}}=50$, $d=7$), it achieves up to a $37\times$ speedup in reaching global faithfulness for Problem \eqref{prob:l1}. Much larger gains are achieved in general, and results on AdaBoost ensembles in \Cref{app:compl_ab} exhibit a similar trend. A finer analysis in \Cref{app:compl_oracle} further shows that, although the constraint-based approach generates more separating samples, it benefits from the efficiency of CP solvers over MILP solvers when no explicit objective function is involved.

\paragraph{Result 2: The active generation of new learners during \textsc{Pace} enhances its pruning capabilities.}
We next show that combining learner generation and pruning via Problem~\eqref{prob:l0} is key to the performance of \textsc{Pace}. To this end, we consider two ablated variants: (i) a version that performs only the learner generation phase, omitting pruning, and (ii) a version that performs only pruning, without augmenting the initial ensemble through learner generation. \Cref{tab:pruning_rf} shows that, across random forest ensembles of varying sizes, \textsc{Pace} consistently outperforms both ablated variants in terms of pruning effectiveness. Neither the learner generation nor the pruning alone matches the performance of the full method. These gains, however, come at the cost of increased computational time. In most cases, all learners in the final ensemble produced by \textsc{Pace} originate from the generation phase, rather than from the initial ensemble. Similar trends are observed in \Cref{app:compl_ab} for AdaBoost ensembles. A comparison to the concurrent pruning-only method \textsc{Fipe}~\citep{emine2025free} is also carried out in \Cref{app:compl_fipel0}, and outlines the superior performance of \textsc{Pace}.

\begin{table}[htbp]
    \centering
    \caption{Statistics of \textsc{Pace} and its ablated variants for random forest ensembles with $n_{\text{est}}~\in~\{25,50,75,100\}$ trees of depth $d=2$ and faithfulness parameters $(\conflevel,\plauslevel)=(0.1,0.1)$. Values include the size $S$ of the ensemble resulting from \textsc{Pace} and its ablated variants, the percentage $P$ of learners that have been newly generated in this ensemble, and the running time $T$ in seconds.}
    \label{tab:pruning_rf}
    \setlength{\tabcolsep}{6pt}
    \resizebox{\textwidth}{!}{%
    \begin{tabular}{ll|rrr|rrr|rrr|rrr}
      \toprule
      & & \multicolumn{3}{c|}{$n_{\text{est}}=25$}  & \multicolumn{3}{c|}{$n_{\text{est}}=50$}  & \multicolumn{3}{c|}{$n_{\text{est}}=75$}  & \multicolumn{3}{c}{$n_{\text{est}}=100$} \\
      \textbf{Dataset} & \textbf{Method} & $S$ & $P$ & $T$ & $S$ & $P$ & $T$ & $S$ & $P$ & $T$ & $S$ & $P$ & $T$ \\
      \midrule
      \rowcolor{gray!15} \textsc{Cancer} & Generate only & 25 & 100\% & 191 & 40 & 100\% & 239 & 49 & 100\% & 339 & 48 & 100\% & 269 \\
      \rowcolor{gray!15} & Pruning only & 7 & -- & 8 & 8 & -- & 11 & 8 & -- & 10 & 8 & -- & 8 \\
      \rowcolor{gray!15} & \textsc{Pace} & 3 & 100\% & 406 & 3 & 100\% & 415 & 3 & 100\% & 385 & 3 & 100\% & 1,238 \\
      \textsc{Compas}  & Generate only & 9 & 100\% & 1 & 4 & 100\% & 1 & 8 & 0\% & 5 & 5 & 100\% & 2 \\
      & Pruning only & 3 & -- & 1 & 2 & -- & 2 & 2 & -- & 1 & 2 & -- & 2 \\
      & \textsc{Pace} & 1 & 100\% & 10 & 2 & 100\% & 16 & 1 & 100\% & 23 & 2 & 50\% & 8 \\
      \rowcolor{gray!15} \textsc{Diabetes} & Generate only & 21 & 100\% & 76 & 25 & 100\% & 61 & 37 & 100\% & 168 & 25 & 100\% & 56 \\
      \rowcolor{gray!15} & Pruning only & 5 & -- & 9 & 4 & -- & 12 & 4 & -- & 7 & 4 & -- & 14 \\
      \rowcolor{gray!15} & \textsc{Pace} & 3 & 100\% & 454 & 3 & 100\% & 669 & 4 & 100\% & 1,288 & 3 & 100\% & 617 \\
      \textsc{Elec2} & Generate only & 5 & 100\% & 4 & 9 & 100\% & 3 & 14 & 100\% & 13 & 6 & 100\% & 1 \\
      & Pruning only & 2 & -- & 2 & 2 & -- & 4 & 3 & -- & 3 & 2 & -- & 2 \\
      & \textsc{Pace} & 2 & 100\% & 19 & 2 & 100\% & 83 & 2 & 100\% & 66 & 2 & 50\% & 6 \\
      \rowcolor{gray!15} \textsc{Fico} & Generate only & 15 & 100\% & 33 & 16 & 100\% & 36 & 14 & 100\% & 100 & 14 & 100\% & 67 \\
      \rowcolor{gray!15} & Pruning only & 3 & -- & 7 & 3 & -- & 13 & 3 & -- & 6 & 3 & -- & 8 \\
      \rowcolor{gray!15} & \textsc{Pace} & 2 & 67\% & 360 & 2 & 100\% & 198 & 2 & 67\% & 297 & 2 & 100\% & 206 \\
      \textsc{House-16H} & Generate only & 25 & 100\% & 12 & 41 & 100\% & 14 & 56 & 100\% & 15 & 62 & 100\% & 10 \\
      & Pruning only & 7 & -- & 23 & 7 & -- & 240 & 7 & -- & 5,137 & 6 & -- & 1,612 \\
      & \textsc{Pace} & 6 & 100\% & 481 & 6 & 100\% & 518 & 6 & 100\% & 1,352 & 6 & 100\% & 4,650 \\
      \rowcolor{gray!15} \textsc{Htru2} & Generate only & 14 & 100\% & 20 & 17 & 100\% & 79 & 15 & 100\% & 84 & 10 & 100\% & 65 \\
      \rowcolor{gray!15} & Pruning only & 4 & -- & 9 & 4 & -- & 7 & 3 & -- & 8 & 4 & -- & 11 \\
      \rowcolor{gray!15} & \textsc{Pace} & 3 & 100\% & 296 & 3 & 100\% & 171 & 3 & 100\% & 90 & 3 & 100\% & 98 \\
      \textsc{Ionosphere} & Generate only & 25 & 100\% & 175 & 48 & 100\% & 389 & 69 & 100\% & 1,112 & 83 & 100\% & 1,428 \\
      & Pruning only & 10 & -- & 8 & 10 & -- & 10 & 9 & -- & 9 & 2 & -- & 9 \\
      & \textsc{Pace} & 2 & 90\% & 1,672 & 3 & 90\% & 16,742 & 3 & 89\% & 3,543 & 3 & 100\% & 26,904\\
      \rowcolor{gray!15} \textsc{Pol} & Generate only & 19 & 100\% & 129 & 27 & 100\% & 146 & 20 & 100\% & 107 & 25 & 100\% & 161 \\
      \rowcolor{gray!15} & Pruning only & 4 & -- & 1 & 4 & -- & 2 & 5 & -- & 2 & 5 & -- & 4 \\
      \rowcolor{gray!15} & \textsc{Pace} & 2 & 100\% & 211 & 2 & 100\% & 888 & 2 & 100\% & 1,934 & 2 & 100\% & 998 \\
      \textsc{Seeds} & Generate only & 25 & 100\% & 316 & 35 & 100\% & 283 & 41 & 100\% & 423 & 51 & 100\% & 498 \\
      & Pruning only & 7 & -- & 24 & 7 & -- & 17 & 7 & -- & 23 & 6 & -- & 21 \\
      & \textsc{Pace} & 2 & 100\% & 415 & 2 & 86\% & 775 & 2 & 100\% & 863 & 2 & 100\% & 454 \\
      \rowcolor{gray!15} \textsc{Spambase} & Generate only & 25 & 100\% & 25 & 45 & 100\% & 156 & 54 & 100\% & 399 & 60 & 100\% & 811 \\
      \rowcolor{gray!15} & Pruning only & 6 & -- & 5 & 6 & -- & 10 & 6 & -- & 28 & 7 & -- & 52 \\
      \rowcolor{gray!15} & \textsc{Pace} & 6 & 100\% & 71 & 4 & 100\% & 439 & 5 & 100\% & 1,991 & 5 & 100\% & 2,314 \\
      \bottomrule
    \end{tabular}
    }
\end{table}

\begin{figure}[htbp]
    \centering
    \begin{subfigure}{\textwidth}
        \centering
        \includegraphics[clip, trim=0.1cm 0.0cm 0.1cm 0.1cm,
                         width=0.33\textwidth]{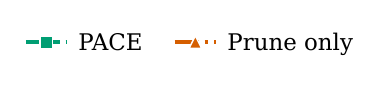}
    \end{subfigure}
    \vspace{-2.4em}

    \begin{subfigure}[t]{0.45\textwidth}
        \centering
        \includegraphics[width=\textwidth]{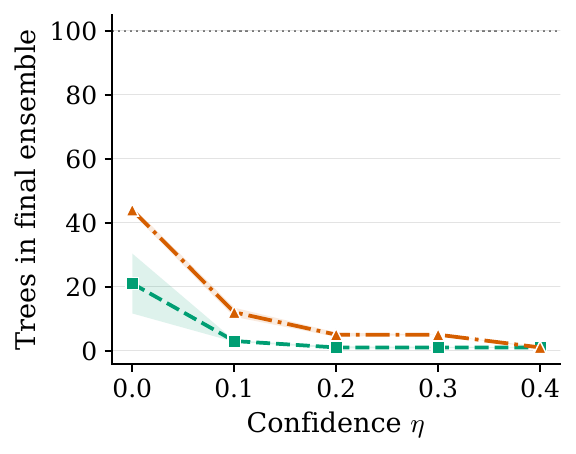}
    \end{subfigure}\hfill
    \begin{subfigure}[t]{0.45\textwidth}
        \centering
        \includegraphics[width=\textwidth]{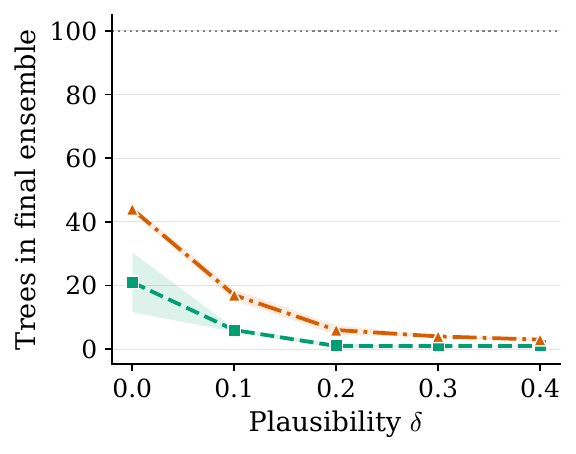}
    \end{subfigure}
    \caption{Compressed ensemble size with varying parameters $\conflevel \geq 0$ and $\plauslevel \geq 0$ for Random Forests ensembles with $n_{\text{est}} = 100$ trees of depth $d = 2$ on the \textsc{Compas} dataset. Left: variation of $\conflevel$ with fixed $\plauslevel = 0$. Right: variation of $\plauslevel$ with fixed $\conflevel = 0$. Shading indicates standard deviation.}
    \label{fig:filter_rf_compas}
\end{figure}

\begin{figure}[htbp]
    \centering
    \begin{subfigure}{\textwidth}
        \centering
        \includegraphics[clip, trim=0.1cm 0.0cm 0.1cm 0.1cm,
                         width=0.33\textwidth]{figures/filter_linechart_d2/legend.pdf}
    \end{subfigure}
    \vspace{-2.4em}

    \begin{subfigure}[t]{0.45\textwidth}
        \centering
        \includegraphics[width=\textwidth]{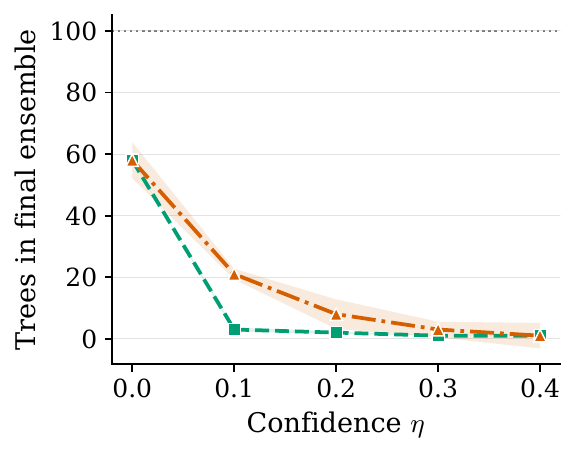}
    \end{subfigure}\hfill
    \begin{subfigure}[t]{0.45\textwidth}
        \centering
        \includegraphics[width=\textwidth]{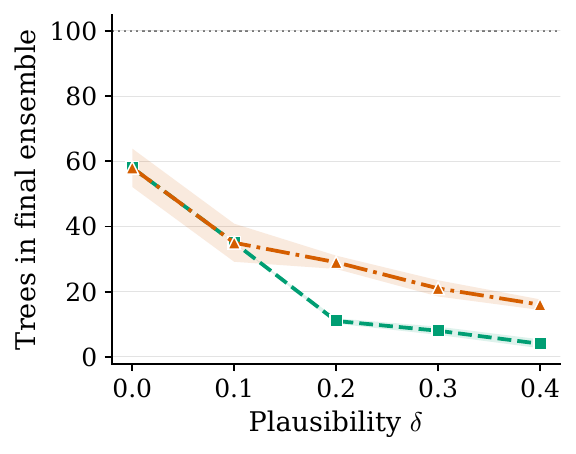}
    \end{subfigure}
    \caption{Compressed ensemble size with varying parameters $\conflevel \geq 0$ and $\plauslevel \geq 0$ for Random Forests ensembles with $n_{\text{est}} = 100$ trees of depth $d = 2$ on the \textsc{Fico} dataset. Left: variation of $\conflevel$ with fixed $\plauslevel = 0$. Right: variation of $\plauslevel$ with fixed $\conflevel = 0$. Shading indicates standard deviation.}
    \label{fig:filter_rf_fico}
\end{figure}

\paragraph{Result 3: A fine control of faithfulness enables additional pruning capabilities in \textsc{Pace}.}
We also examine how the confidence- and distribution-aware faithfulness region introduced in \Cref{sec:inner} affects compression capabilities. Using random forest ensembles with $n_{\text{est}}=100$ trees of depth $d=2$, we vary $\conflevel\geq 0$ and $\plauslevel\geq 0$\footnote{We rescale the parameter $\plauslevel$ so that it reflects the proportion of outlier detected by the isolation forest. With $\plauslevel=0$, all samples are flagged as inliers, while with $\plauslevel=1$, all samples are flagged as outliers.} and report the size of the ensemble returned by \textsc{Pace} for the \textsc{Compas} and \textsc{Fico} datasets, alongside that of the \say{pruning only} variant described previously.
\Cref{fig:filter_rf_compas,fig:filter_rf_fico} show that controlling faithfulness can drastically improve pruning \textsc{Pace}'s capabilities: on \textsc{Fico}, using $\conflevel=0.1$ reduces the ensemble from $58$ to $3$ trees compared to $\conflevel=0$, and a similar reduction is observed on \textsc{Compas}. The benefit compounds with active generation, although in a dataset-dependent way. On \textsc{Compas}, generation helps even in the strict regime with $\conflevel=\plauslevel=0$, where \textsc{Pace} returns about half as many trees as its prune-only counterpart. On \textsc{Fico}, by contrast, the two variants are essentially indistinguishable at $\conflevel=\plauslevel=0$, and a gap only emerges once faithfulness is relaxed. At $\plauslevel=0.4$, the prune-only variant still retains $18$ trees against $4$ for \textsc{Pace}. This suggests that active generation and faithfulness relaxation are complementary; tighter faithfulness regions open up compression opportunities that pruning of the initial ensemble $\initlearners$ alone cannot exploit. Analogous trends on AdaBoost ensembles for the same datasets are reported in Appendix~\ref{app:compl_ab}.


%% file: sections/conclusions.tex
\section{Conclusion}
\label{sec:conclusions}

This paper introduced \textsc{Pace}, a framework that unifies compression and pruning techniques to provide a compact and faithful representation of trained ensemble models. In contrast to prior methodologies from the literature, the proposed method combines the active generation of new learners with a strong pruning operation based on the $\ell_0$-norm. In addition, \textsc{Pace} incorporates a flexible notion of faithfulness, allowing one to range from exact replication of the initial ensemble’s predictions to more relaxed requirements to obtain enhanced pruning capabilities. Numerical results demonstrate that \textsc{Pace} consistently achieves more effective pruning and compression than existing approaches, and that further improvements can be obtained through its finer control of faithfulness enforcement.

\noindent\textbf{Limitations.}
A first limitation of \textsc{Pace} lies in its computational cost, which is higher than that of previously proposed pruning-only or compression-only methods. This overhead is primarily driven by the combinatorial nature of the $\ell_0$-norm pruning problem solved after the active generation of new learners. Nevertheless, we have observed that the flexible control of faithfulness can help mitigate this cost. Second, the current implementation of \textsc{Pace} is restricted to ensembles composed of tree-based learners. However, the proposed strategy could be naturally extended to other families of learners, provided that an appropriate mathematical model of their prediction functions can be established. Finally, the proposed characterization of faithfulness in \textsc{Pace} could be further enriched by incorporating additional considerations into the generation of separating samples. Such extensions could be readily integrated into the current implementation, provided that appropriate mathematical models of these new considerations are available.

\noindent\textbf{Perspectives.}
This work opens several directions for future research. A first promising avenue is the study of methods for generating improving learners during the first phase of \textsc{Pace}. In particular, it would be valuable to investigate the impact of solving the underlying pricing problem either exactly or heuristically. Prior work on optimal decision tree design could be leveraged to explore this direction. Another way to further enhance \textsc{Pace} would be to refine the notion of ensemble complexity beyond simply counting the number of base learners. For instance, for tree-based ensembles, one could instead consider the total number of nodes or leaves the trees, which more accurately reflects the memory footprint of the learner. This notion could be readily incorporated into the current framework through an appropriate scaling of the learners’ weights. Finally, a natural extension of \textsc{Pace} would be to generate improving learners directly with respect to the $\ell_0$-norm pruning objective, rather than relying on its $\ell_1$-norm surrogate. In this context, branch-and-cut-and-price techniques, generalizing column generation to mixed-integer linear programs, could be adapted for this purpose. Although such an approach would likely increase the computational burden of the method, it would provide an optimal strategy for compressing a trained ensemble model with faithfulness guarantees.

%% file: sections/supp-literature.tex
\section{Related Works}
\label{sec:related_works}

This appendix provides a review of related works on ensemble training, pruning, and compression techniques, as well as connections to other machine learning models beyond ensemble models.

\subsection{Ensemble Model Training}
\label{sec:related_works_lp_boosting}

Training an ensemble model amounts to building a collection of learners and weights using a given dataset. Historically, bagging techniques \citep{breiman1996bagging} were proposed for this task, in which learners are trained simultaneously on variations of the training data generated via bootstrap sampling to reduce prediction variance \citep{hesterberg2011bootstrap}. This paradigm has been largely replaced by more recent boosting methods, in which learners are built sequentially with a focus on reducing prediction error~\citep {freund1996experiments}. Two landmark algorithms in this vein are AdaBoost \citep{schapire2013explaining} and Extreme Gradient Boosting \citep{friedman2000additive}, the latter being most commonly known as XGBoost. Both interpret the boosting task as functional gradient descent in the predictor space, pointing towards new learners that reduce prediction errors~\citep{friedman2001greedy}.

While these strategies apply to abstract learners, particular effort has been devoted to tree-based learners, leading to random forest ensemble models \citep{rigatti2017random}. Their structure can be leveraged within training methods to obtain strong aggregated predictors. For instance, combining bootstrap resampling with randomized feature selection at each tree split has been shown to decorrelate trees and reduce variance~\citep{breiman2001random}. Related methods further increase randomization by selecting split thresholds at random, often improving computational efficiency and robustness \citep{geurts2006extremely}. 

Another ensemble training technique comes from the linear programming literature, where the training task has been identified as a column generation procedure \citep{Demiriz2002}. With this approach, training a new learner can be viewed as solving a pricing problem \citep{warmuth2008entropy}, focusing on reducing the prediction error of previously generated learners. The parametrization of this pricing problem has led to different variants of LP-boosting methods, focusing on margin maximization \citep{grove1998boosting}, nominal improvement from Lagrange duals \citep{shen2010dual}, or gradient values \citep{combettes2020boosting}. A unified view of these methods is provided in \citep{akkerman2025boosting}.

\subsection{Ensemble Model Pruning}

Training methods for ensemble models typically generate a large number of learners that may be redundant. Various studies have proposed pruning techniques to reduce the number of learners while maintaining, or even improving, generalization performance. Early approaches relied on ordering and ranking strategies, in which individual learners are scored based on standalone accuracy or diversity, and only the top candidates are retained \citep{caruana2006getting,martinez2007using}. These methods can also be seen as a subset selection problem, explicitly searching for a collection of models whose joint predictions maximize a validation criterion \citep{martinez2008analysis}. Practical approaches rely on greedy forward-backward selection, local search, or metaheuristics \citep{partalas2009pruning}.  More recently, optimization-based pruning methods have been proposed, trading this discrete subset selection problem for continuous weighting schemes. Typical examples include $\ell_1$-penalized weighting \citep{li2012diversity}, convex margin maximization \citep{zhang2006ensemble}, or constrained optimization formulations that jointly optimize predictive accuracy and ensemble size \citep{qian2015pareto}. Recently, studies have investigated the possibility of faithful ensemble pruning. While some have focused on achieving a suitable faithful-pruning tradeoff \citep{amee2023novel,liu2023forestprune}, others rather focus on exact faithfulness guarantees via a separation problem \citep{emine2025free}.

\subsection{Ensemble Model Compression}
\label{sec:related_works_ensemble_compression}

Beyond pruning, a complementary line of work focuses on compression techniques, aiming to generate a new, compact ensemble model from scratch that approximates the behavior of a larger one. In particular, a few works have attempted to compress existing tree ensembles into a single decision tree. Some approximate approaches train a surrogate tree from synthetic data obtained by sampling pseudo-covariates and labeling them with predictions from the original ensemble, using a custom split-selection procedure to stabilize the resulting tree \citep{zhou2016interpreting}. While they can yield substantial compression, they do not guarantee exact faithfulness to the original model. In contrast, a dynamic-programming algorithm can be devised to construct a single decision tree that is certifiably faithful to the original ensemble \citep{vidal2020born}. While this guarantees exact faithfulness, the objective is no longer to learn a simplified surrogate, but to recover an exact representation, which may remain significantly large in practice.

\subsection{Beyond Ensemble Models}

Apart from ensembles of learners, pruning and compression techniques have been proposed for more general machine learning models. Recently, a significant amount of research has focused on techniques to reduce the memory footprint of deep neural networks via impact-based \citep{dong2019network,liebenwein2021lost}, optimization-based \citep{benbaki2023fast,zhang2022advancing}, quantization-based \citep{kuzmin2023pruning}, or structural-based \citep{meng2024falcon} methods. These approaches can be mainly categorized into unstructured pruning, where individual weights are removed based on their importance, and structured pruning, where entire neurons or filters are removed \citep{liu2019rethinking}. We refer to \citep{cheng2024survey} for a broader overview of these connected works.

%% file: sections/supp-cg.tex
\section{Technical Specifications on the Learner Generation}
\label{app:proof-pricing}

This appendix presents the main ingredients from the column generation literature relevant to the creation of new learners in \textsc{Pace}, along with a proof of \Cref{prop:pricing}. We refer the reader to \citep[Chapiter 2]{desrosiers2026column} for a complete description of these techniques.

\subsection{Column Generation Ingredients}

Column generation is a technique for solving linear programs of the form
\begin{equation}
    \label{prob:lp-extended}
    \left\{
    \begin{array}{lrl}
        \min & \mathbf{d}^\top \mathbf{w} \\
        \text{s.t.} & \mathbf{A}\mathbf{w} &\geq \mathbf{b} \\
        & \mathbf{w} &\geq \mathbf{0}
        .
    \end{array}  
    \right.
\end{equation}
Instead of directly addressing this formulation with the entire variable $\mathbf{w} \in \R^n$, column generation considers a restricted version of the problem defined as
\begin{equation}
    \label{prob:lp-master}
    \left\{
    \begin{array}{lrl}
        \min & \mathbf{d}_I^\top \mathbf{w}_{I} \\
        \text{s.t.} & \mathbf{A}_{I}\mathbf{w}_{I} &\geq \mathbf{b} \\
        & \mathbf{w}_{I} &\geq \mathbf{0}
    \end{array}  
    \right.
\end{equation}
over a subset $I \subseteq \{1,\ldots,n\}$ of the variable indices, where $\mathbf{d}_{I}$ and $\mathbf{w}_{I}$ (resp. $\mathbf{A}_{I}$) denote the restriction to the indices (resp. columns) indexed by $I$. The column generation procedure iteratively solves the restricted problem \eqref{prob:lp-master}, and seeks a new index allowing to improve its optimal value when added to $I$. The operation is carried out by solving a so-called pricing problem of the form
\begin{equation}
    \label{prob:lp-pricing}
    \textstyle
    c^\star = \min_{i \in \{1,\dots,n\}} \ d_i - \mathbf{a}_i^\top \mathbf{v}^{\star}
    ,
\end{equation}
where $\mathbf{v}^{\star} \in \mathbb{R}^m$ denotes the optimal solution to the dual of problem \eqref{prob:lp-master}, given by
\begin{equation}
    \label{prob:lp-dual}
    \left\{
    \begin{array}{lrl}
        \max & \mathbf{b}^\top \mathbf{v} \\
        \text{s.t.} & \mathbf{A}^{\top}_{I}\mathbf{v} &\geq \mathbf{d}_I \\
        & \mathbf{v} &\geq \mathbf{0}
        .
    \end{array}  
    \right.
\end{equation}
The quantity $c_i = d_i - \mathbf{a}_i^\top \mathbf{v}^{\star}$ is referred to as \emph{reduced cost} of the index $i$, and represents the marginal improvement in the optimal value of the restricted problem \eqref{prob:lp-master} when this index is added to the subset $I$. We have $c_i=0$ whenever $i \in I$.
If $c^\star < 0$, then the index that achieves the minimum value in \eqref{prob:lp-pricing} is added to $I$ to improve the optimal value of the restricted problem \eqref{prob:lp-master}, and the operation is repeated. Ultimately, one obtains $c^\star = 0$, and no more indices can improve this optimal value. At this point, the solution to problem \eqref{prob:lp-extended} can be retrieved from that of its restriction \eqref{prob:lp-master}, and the column generation procedure terminates. This can occur even (and hopefully) if not all indices have been included in $I$.

\subsection{Proof of \Cref{prop:pricing}}

Similarly to \eqref{prob:l1-lp}, problem \eqref{prob:l1-ext} can be expressed by instantiating the linear program~\eqref{prob:lp-extended} using $\mathbf{d} = \1$ and $\mathbf{b} = \1$, together with a matrix $\mathbf{A}$ with $\card{\learnerset}$ columns and $\card{\initsamplespace} \cdot (\card{\predictset}-1)$ rows, where the entry corresponding to some $\learner{} \in \learnerset$ and $(\sample{},\predict{}) \in \initsamplespace \times \predictset$ is defined as
\begin{equation}
    \mathbf{A}_{\learner{},(\sample{},\predict{})} = \learner{\initpredict}(\sample{}) -
    \learner{\predict{}}(\sample{})
\end{equation}
with $\initpredict = \majorityvote_{\initlearners}(\initlearnerweights,\sample{})$. Note that the optimization variable in this problem is the vector of the base learner weights only, while the sample subset $\initsamplespace$, the label classes $\predictset$, and the set of base learners $\learnerset$ are fixed. In a similar fashion to how problem \eqref{prob:lp-master} is a restricted version of \eqref{prob:lp-extended}, problem \eqref{prob:l1-lp} is a restricted version of \eqref{prob:l1-ext} where only a subset of base learners in the whole family $\learnerset$ is involved. It involves a reduced version of the vectors $\mathbf{d}$ and $\mathbf{w}$ (resp. matrix $\mathbf{A}$), where only the entries (resp. columns) of the base learners in the current ensemble $\learners$ are considered. Translating \eqref{prob:lp-pricing} to our setting, we obtain the pricing problem
 \begin{equation}
    c^\star = \min_{\learner{} \in \learnerset} \ 1 - \sum_{\sample{} \in \samplespace}
    \sum_{\predict{} \neq \initpredict}
    \sampleweight{\sample{},\predict{}} \cdot 
    \big(
        \learner{\initpredict}(\sample{}) -
        \learner{\predict{}}(\sample{})
    \big)
\end{equation}
by expanding the definitions of $\mathbf{d}$ and $\mathbf{A}$ as in \Cref{prop:pricing}. The conclusion on whether there exists an improving base learner is then obtained from the quantity $c^\star$ as in the column generation procedure.

%% file: sections/supp-oracle.tex
\section{Technical Specifications on the Separation Problem}
\label{appendix:oracle}

This appendix provides supplementary details on the formulation used in \textsc{Pace} to generate separating samples while solving Problems \eqref{prob:l1} and \eqref{prob:l0}. A complete formulation of the separation problem is also provided, gathering all the ingredients described in \Cref{sec:inner}.

\subsection{Plausibility Score Computation}
\label{appendix:oracle:plausibility}

The plausibility score is modeled in \textsc{Pace} through an isolation forest $\mathcal{T}$ as $A(\sample{})=\sum_{\learner{} \in \mathcal{T}} b_{\learner{}}(\sample{})$ where $b_{\learner{}}(\sample{}) \geq 0$ denotes the path length of a sample $\sample{} \in \samplespace$ through isolation tree~$\learner{}$, i.e., the depth of the leaf reached, corrected using an adjustment factor based on the support of this leaf. Specifically, denote by $d_{\learner{}}(\sample{}) \in \N$ the depth of the leaf reached by the sample and by $s_{\learner{}}(\sample{}) \in \N$ the number of training samples that reached this leaf during the isolation forest building. Then, this corrected path length is given by
\begin{equation}
    b_{\learner{}}(\sample{}) \;=\;
    \begin{cases}
    d_{\learner{}}(\sample{}) & s_{\learner{}}(\sample{}) \le 1 \\
    d_{\learner{}}(\sample{}) + c(s_{\learner{}}(\sample{})), & s_{\learner{}}(\sample{}) > 1
    \end{cases}
\end{equation}
where $c: \N \to \R_+$ is the expected additional path length that would have been necessary to further isolate the sample $\sample{} \in \samplespace$ from the other training samples that reached the leaf. In practice, isolation forests express this quantity as
\begin{equation}
    c(s) \;=\;
    \begin{cases}
        0, & s \le 1\\
        1, & s = 2\\
        2(\ln(s-1) + \gamma) - \tfrac{2(s-1)}{s}, & s > 2
    \end{cases}
\end{equation}
where $\gamma \approx 0.57721$ is the Euler-Mascheroni constant. Note that a scaled version 
\begin{equation}
    \textstyle
    \samplespace_{\plauslevel_{\text{scaled}}}=\big\{
        \sample{} \in \samplespace
        \mid
        2^{-A(\sample{}) / c_\text{max}} \leq \plauslevel_{\text{scaled}}
    \big\}
\end{equation}
of the plausibility region with some threshold $\plauslevel_{\text{scaled}} \in [0,1]$ is sometimes used in the literature where $c_{\max} = c(s_{\max})$ accounts for the maximum number of training samples $s_{\max} \in \N$ used to build each tree in the isolation forest. A simple transformation allows to recover our formulation since
\begin{equation}
    2^{-A(\sample{}) / c(s_\text{max})} \leq \plauslevel_{\text{scaled}} \iff A(\sample{}) \geq \plauslevel
\end{equation}
with $\plauslevel = -c(s_\text{max}) \cdot \log_2(\plauslevel_{\text{scaled}})$.

\subsection{Encoding Continuous Features}
\label{appendix:oracle:encoding-features}

Each feature of a sample $\sample{} \in \samplespace$ may take values in a continuous domain, i.e., $x_i \in \R$. Although CP solvers can only handle discrete variables, this technical limitation can be readily overcome. Indeed, tree-based learners only contain finitely many nodes, and therefore only involve finitely many split levels $\samplespace_i \subseteq \R$ for the feature $x_i \in \R$. Consequently, one can pre-compute, for every feature $i \in \{1,\dots,\sampledim\}$, the finite set of possible split levels $\samplespace_i$ induced by the initial ensemble $(\initlearners,\initlearnerweights)$ given to \textsc{Pace}. A discrete variable $k_i \in \{1,\dots,\card{\samplespace_i}\}$ can then be introduced to encode the interval between consecutive elements of $\samplespace_i$ in which the feature value lies. In the special case of binary features, we have $\samplespace_i = \{0,1\}$, while for categorical features, the number of split levels matches the number of categories. With this encoding, the CP solver becomes fully agnostic to the actual value of continuous features, and instead bases its search for separating points on the intervals defined by the split levels of the initial ensemble $(\initlearners,\initlearnerweights)$ and of the isolation forest $\mathcal{T}$. This induces no loss of accuracy, since any feature value $x_i \in [a,a')$ lying between two consecutive split levels $(a,a')$ of $\samplespace_i$ belongs to the same decision region of the tree, and is therefore indistinguishable from its perspective.

\subsection{Encoding Continuous Coefficients}
\label{appendix:oracle:encoding-coefficients}

In a similar fashion to the encoding of continuous feature values, a continuous value can be taken by the entries of weights $(\initlearnerweights,\learnerweights)$ from the initial and current ensemble involved in \textsc{Pace}, as well as by coefficients $c(s)$ appearing in the definition of the plausibility score computation discussed in \Cref{appendix:oracle:plausibility}. Although they appear as constants in the separation problem, CP solvers only support discrete coefficients. These coefficients are used in the expression of the per-label confidence scores, which are themselves decision variables and must therefore be represented as integer-valued quantities. To overcome this issue, we rescale any continuous coefficient $\alpha \in \R$ appearing in the separation problem as
\begin{equation}
    \alpha_{\text{scaled}} = \operatorname{round}\!\left(\alpha \cdot 10^d\right).
\end{equation}
where $d \in \N$ determines the decimal precision: the coefficient is first scaled by $10^d$ and then rounded to the nearest integer. By default, \textsc{Pace} uses $d=9$. Note that the separation problem is invariant with respect to positive rescaling of the weights $(\initlearnerweights,\learnerweights)$ and of the coefficients $c(s)$. Hence, this is done without loss of generality, provided that a similar rescaling of the parameters $\conflevel$ and $\plauslevel$ involved in the definition of the regions $\samplespace_{\conflevel}$ and $\samplespace_{\plauslevel}$ is performed. Note that such scaling has usually little impact on the numerical efficiency and stability of CP solvers used to tackle the separation problem.

\subsection{Complete Separation Problem Formulation}

With all the ingredients introduced in \Cref{sec:inner}, the complete formulation of the separation problem used in \textsc{Pace} for some pair $(\initpredict{},\predict{}) \in \predictset \times \predictset$ of distinct labels reads
\begin{equation}
    \left\{
    \begin{array}{rll}
        \sum_{\learner{} \in \initlearners} \initlearnerweight{\learner{}} (\learner{\initpredict}(\sample{})-\learner{\tilde{\predict{}}}(\sample{})) > \conflevel & \forall \tilde{\predict{}} \neq \initpredict
        \\
        \sum_{\learner{} \in \learners\phantom{{}^o}} \learnerweight{\learner{}} (\learner{\predict{}}(\sample{})-\learner{\initpredict}(\sample{})) > 0 \\
        \sum_{\learner{} \in \mathcal{T}} b_{\learner{}}(\sample{}) \geq \plauslevel 
        \\
        \sample{} \in \samplespace
    \end{array}
    \right.
\end{equation}
for fixed parameters $\conflevel \geq 0$ and $\plauslevel \geq 0$. Upon the model decision tree learner prediction functions as described by \eqref{oracle:base}, the extended and practical formulation of this separation problem reads
\begin{equation}
    \label{prob:full-cp-model}
    \setlength{\arraycolsep}{2pt}
    \left\{
    \begin{array}{lll}
        \sum_{\learner{} \in \initlearners} \initlearnerweight{\learner{}} (r_{\learner{}\initpredict}-r_{\learner{}\tilde{\predict{}}}) &> \conflevel & \forall \tilde{\predict{}} \neq \initpredict
        \\
        \sum_{\learner{} \in \learners\phantom{{}^o}} \learnerweight{\learner{}} (r_{\learner{}\predict{}}-r_{\learner{}\initpredict}) &> 0 \\
        \sum_{\learner{} \in \mathcal{T}}
        \sum_{v \in \mathcal{V}_{\learner{}}}
        b_{\learner{}v} \cdot z_{\learner{}v} &\geq \plauslevel
        \\
        \sum_{v \in \mathcal{V}_{\learner{}}}v_{\predict{}} \cdot z_{\learner{}v} - r_{\learner{}\predict{}} & = 0 & \forall \learner{} \in \initlearners \cup \learners \cup \mathcal{T}
        \\
        \textstyle
        \sum_{v \in \mathcal{V}_{\learner{}}} z_{\learner{}v} &= 1 & \forall \learner{} \in \initlearners \cup \learners \cup \mathcal{T}
        \\
        \multicolumn{2}{l}{z_{\learner{}v}=1 \textstyle\implies \big(\bigwedge_{(i,a)\in\Gamma_{\learner{}v}^{+}} x_i>a\big)\land\big(\bigwedge_{(i,a)\in\Gamma_{\learner{}v}^{-}} x_i\leq a\big)} & \forall \learner{} \in \initlearners \cup \learners \cup \mathcal{T}, \ \forall v \in \mathcal{V}_{\learner{}}
        \\
        \multicolumn{1}{r}{r_{\learner{}\predict{}}} &\in\N & \forall \learner{} \in \initlearners \cup \learners \cup \mathcal{T}, \ \forall v \in \mathcal{V}_{\learner{}}
        \\
        \multicolumn{1}{r}{z_{\learner{}v}} &\in\{0,1\} & \forall \learner{} \in \initlearners \cup \learners \cup \mathcal{T}, \ \forall v \in \mathcal{V}_{\learner{}}
        \\
        \multicolumn{1}{r}{x_i} &\in \samplespace_i & \forall i \in \{1,\dots,\sampledim\}
    \end{array}
    \right.
\end{equation}
for a fixed pair $(\initpredict{},\predict{}) \in \predictset \times \predictset$ of distinct prediction labels. In this formulation, the discrete sets $\{\samplespace_1,\dots,\samplespace_\sampledim\}$ are obtained in pre-processing as described in \Cref{appendix:oracle:encoding-features} from the split levels appearing in the original ensemble $(\initlearners,\initlearnerweights)$. Similarly, all coefficients $(r_{\learner{}v},v_{\predict{}},b_{\learner{}v})$ and sets \smash{$(\Gamma_{\learner{}v}^{-},\Gamma_{\learner{}v}^{+})$} are obtained in pre-processing from the structure of the initial ensemble $\initlearners$, the current ensemble $\learners$, and the isolation forest $\mathcal{T}$. Continuous coefficients are rescaled as described in \Cref{appendix:oracle:encoding-coefficients} if necessary. Any addition of a new improving learner $\learner{} \in \learnerset$ to the current ensemble $\learners$ during \textsc{Pace} can be done online by adding the corresponding sets of variables $\{z_{\learner{}v}\}_{v \in \mathcal{V}_{\learner{}}}$ and $\{r_{\learner{}\predict{}}\}_{v \in \mathcal{V}_{\learner{}}}$ together with their associated constraints in the model, without re-creating it from scratch.

%% file: sections/supp-datasets.tex
\section{Experimental Setup and Datasets}
\label{app:datasets}

\textsc{Pace} is written in Python and is available under MIT license. An anonymized version of the code is attached to the paper submission, and a public version will be released Github and PyPI upon acceptance decision. Both pruning Problems \eqref{prob:l0} and \eqref{prob:l1} are solved using Gurobi version 10.0~\citep{gurobi}, while the separation problem described in \Cref{sec:inner} is solved using the OR-Tools CP-SAT version 9.14~\citep{cpsatlp}. Moreover, we use Scikit-learn \citep{scikit-learn} to generate improving learners as described in \Cref{sec:outer}. All experiments are run on eight threads of a compute node equipped with AMD Genoa 9654 @2.4GHz cores, along with 2 GB of memory per core. Numerical results provided are averaged over 5 independent runs with a 20-hour time limit.

The datasets considered in our setup are commonly used to benchmark binary or multi-class classification tasks \citep{akkerman2025boosting,emine2025free}. They are openly accessible and provided with our code for reproducibility purpose. \Cref{tab:datasets} displays their number of binary features, numerical features, samples, classes, and the distribution of these classes. In pre-processing, we discretize continuous features into 10 bins based on quantiles value. If at least one bin appears to be empty, we instead use uniformly-spaced bins based. We also perform a $80\%/20\%$ train-test split of the dataset, and use the training part to initialize the faithfulness region $\initsamplespace \subseteq \samplespace$ in \textsc{Pace}, filtering-out any sample lying outside of the regions $\samplespace_{\conflevel}$ and $\samplespace_{\plauslevel}$. This pre-processing operation is also part of the code accompanying this paper.

\begin{description}[leftmargin=0.5cm]
    \item \textsc{Cancer} is a dataset sourced from the Institute of Oncology in Ljubljana for forecasting recurrence events of breast cancer in Wisconsin \citep{breast_cancer_14}.
    \item \textsc{Compas} is a dataset for predicting a criminal defendant's likelihood of re-offending in Broward County, Florida \citep{angwin2016machine}.   
    \item \textsc{Diabetes} is a dataset from the National Institute of Diabetes and Digestive and Kidney Disease for predicting diabetes based on diagnostic measurements \citep{pima_diabetes}.
    \item \textsc{Elec2} is a dataset for predicting the increase or decrease of electricity prices in Australian New South Wales \citep{harries1999splice}. 
    \item \textsc{Fico} is a credit risk assessment dataset for predicting default risk for an explainable machine learning challenge \citep{fico}.
    \item \textsc{House-16H} is a dataset from the 1990 United States Census for predicting house prices, split into two buckets \citep{vanschoren2014openml}.
    \item \textsc{Htru2} is a dataset for classifying neutron stars as genuine pulsar or non-pulsar, supporting the discovery of new spatial systems \citep{htru2_372}.
    \item \textsc{Ionosphere} is a dataset for detecting the presence of a specific structure of free electrons in the ionosphere based on high-frequency signals from radars located in Goose Bay, Labrador \citep{sigillito1989classification}.
    \item \textsc{Pol} is a dataset derived from commercial telecommunication applications for predicting a target variable split into two buckets \citep{vanschoren2014openml,weiss1995rulebasedmachinelearningmethods}.
    \item \textsc{Seeds} is a dataset for characterizing geometrical properties of three varieties of wheat kernels from high-quality X-ray images \citep{seeds_236}.
    \item \textsc{Spambase} is a dataset for classifying emails as spam based on words and frequency measurement of characters \citep{spambase_94}.
\end{description}

\begin{table}[hbtp]
\centering
\caption{Datasets characteristics.}\label{tab:datasets}
\begin{tabular}{l|rrrrr}
\toprule
\textbf{Dataset}  & \multicolumn{1}{c}{\textbf{Binary feat.}} & \multicolumn{1}{c}{\textbf{Numerical feat.}} & \multicolumn{1}{c}{\textbf{Samples}} & \textbf{Classes} & \textbf{Class distribution} \\
\midrule
\rowcolor{gray!15}\textsc{Cancer} & 0 & 8 & 683 & 2 & 0.650~/~0.350 \\
\textsc{Compas} & 12 & 0 & 6,907 & 2 & 0.537~/~0.463 \\
\rowcolor{gray!15}\textsc{Diabetes} & 0 & 8 & 768 & 2 & 0.652~/~0.348 \\
\textsc{Elec2} & 0 & 7 & 38,474 & 2 & 0.500~/~0.500 \\
\rowcolor{gray!15}\textsc{Fico} & 17 & 0 & 10,459 & 2 & 0.522~/~0.478 \\
\textsc{House-16H} & 0 & 16 & 13,488 & 2 & 0.500~/~0.500  \\
\rowcolor{gray!15}\textsc{Htru2} & 0 & 8 & 17,898 & 2 & 0.908~/~0.092 \\
\textsc{Ionosphere} & 1 & 32 & 351 & 2 & 0.640~/~0.360 \\
\rowcolor{gray!15}\textsc{Pol} & 0 & 26 & 10,082 & 2 & 0.500~/~0.500 \\
\textsc{Seeds} & 0 & 7 & 210 & 3 & 0.330~/~0.335~/~0.335 \\
\rowcolor{gray!15}\textsc{Spambase} & 0 & 57 & 4,601 & 2 & 0.606~/~0.394 \\
\bottomrule
\end{tabular}
\end{table}

\FloatBarrier

%% file: sections/supp-results.tex
\section{Complementary Results}
\label{app:comlementary_results}

This appendix gathers complementary results not included in the main text, including experiments on AdaBoost ensembles and further analyses of \textsc{Pace} 's internal behavior.

\subsection{Numerical Analyses with AdaBoost Ensembles}
\label{app:compl_ab}

\begin{figure}[htbp]
    \centering

    \begin{subfigure}{\textwidth}
        \centering
        \includegraphics[clip, trim=0.1cm 0.1cm 0.1cm 0.1cm,
                         width=0.75\textwidth]{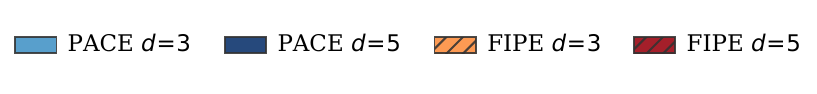}
    \end{subfigure}
    \vspace{-2.4em}

    \begin{subfigure}[t]{0.99\textwidth}
        \centering
        \includegraphics[width=\textwidth]{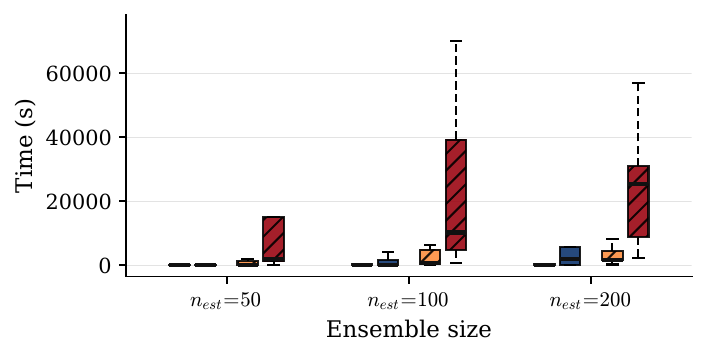}
    \end{subfigure}\hfill
    \caption{Time to achieve global faithfulness for problem~\eqref{prob:l1} via the constraint-based formulation in \textsc{Pace} and the constraint-based one in \textsc{Fipe} for retrieving separating samples. Boxes show AdaBoost ensembles of size $n_{\text{est}} \in \{50, 100, 200\}$, with color shade encoding tree depth $d \in \{5, 7\}$. For a fair comparison, results are aggregated on the datasets listed in \Cref{app:datasets} (11 in total) where both \textsc{Fipe} and \textsc{Pace} terminate within the 20-hour time limit, i.e. on 9 data sets for $n_{\text{est}}=50$, 8 datasets for $n_{\text{est}}=100$, and 5 datasets for $n_{\text{est}}=200$. In practice, \textsc{Pace} terminated over a larger number of datasets, namely {11} for $n_{\text{est}}=50$, {11} for $n_{\text{est}}=100$, and {10} for $n_{\text{est}}=200$.}
    \label{fig:compu_time_ada}
\end{figure}

\begin{figure}[hbtp]
    \centering

    \begin{subfigure}{\textwidth}
        \centering
        \includegraphics[clip, trim=0.1cm 0.1cm 0.1cm 0.1cm,
                         width=0.4\textwidth]{figures/filter_linechart_d2/legend.pdf}
    \end{subfigure}
    \vspace{-2.4em}

    \begin{subfigure}[t]{0.49\textwidth}
        \centering
        \includegraphics[width=\textwidth]{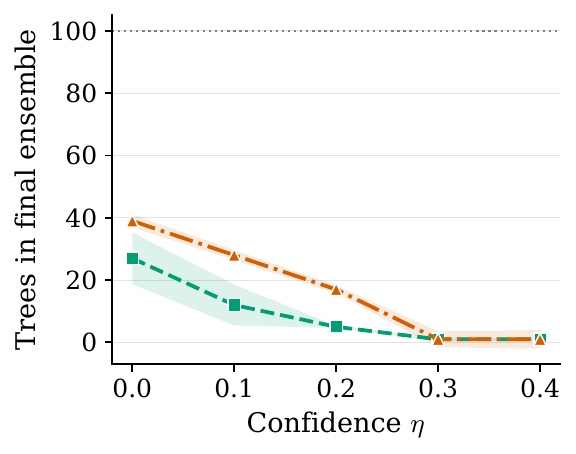}
    \end{subfigure}\hfill
    \begin{subfigure}[t]{0.49\textwidth}
        \centering
        \includegraphics[width=\textwidth]{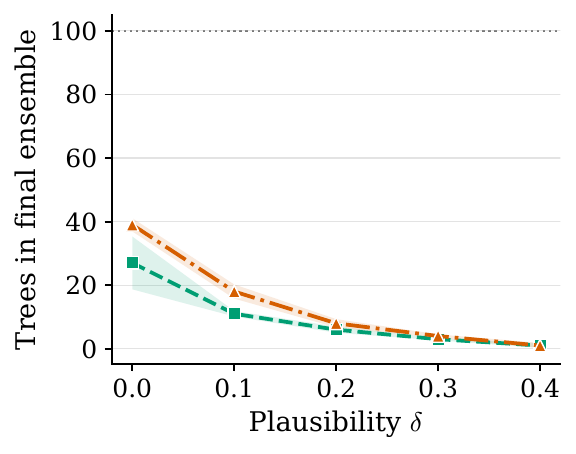}
    \end{subfigure}
    \caption{Compressed ensemble size with varying parameters $\conflevel \geq 0$ and $\plauslevel \geq 0$ for AdaBoost ensembles with $n_{\text{est}} = 100$ trees of depth $d = 2$ on the \textsc{Compas} dataset. Left: variation of $\conflevel$ with fixed $\plauslevel = 0$. Right: variation of $\plauslevel$ with fixed $\conflevel = 0$. Shading indicates standard deviation.}
    \label{fig:filter_ab_compas}
\end{figure}

\begin{figure}[hbtp]
    \centering

    \begin{subfigure}{\textwidth}
        \centering
        \includegraphics[clip, trim=0.1cm 0.1cm 0.1cm 0.1cm,
                         width=0.4\textwidth]{figures/filter_linechart_d2/legend.pdf}
    \end{subfigure}
    \vspace{-2.4em}

    \begin{subfigure}[t]{0.49\textwidth}
        \centering
        \includegraphics[width=\textwidth]{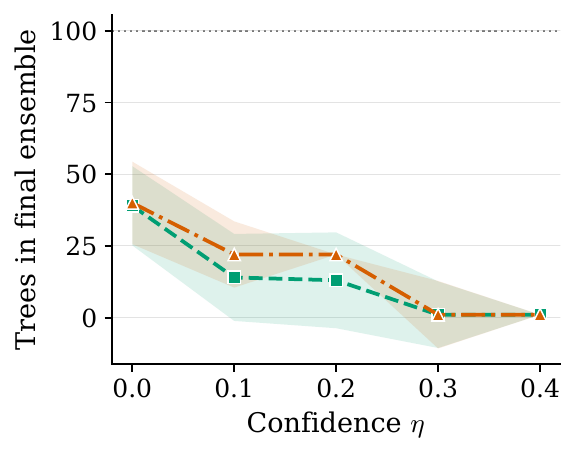}
    \end{subfigure}\hfill
    \begin{subfigure}[t]{0.49\textwidth}
        \centering
        \includegraphics[width=\textwidth]{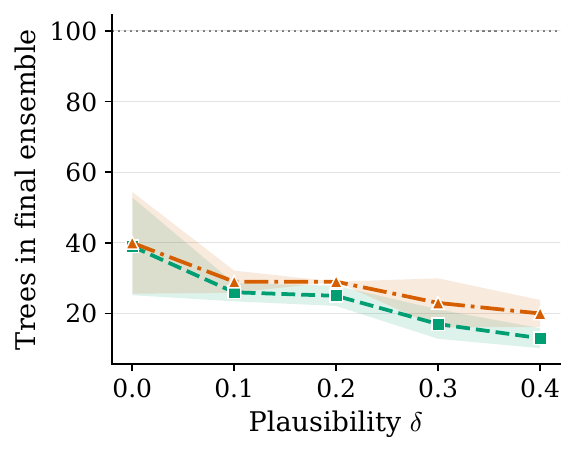}
    \end{subfigure}
    \caption{Compressed ensemble size with varying parameters $\conflevel \geq 0$ and $\plauslevel \geq 0$ for AdaBoost ensembles with $n_{\text{est}} = 100$ trees of depth $d = 2$ on the \textsc{Fico} dataset. Left: variation of $\conflevel$ with fixed $\plauslevel = 0$. Right: variation of $\plauslevel$ with fixed $\conflevel = 0$. Shading indicates standard deviation.}
    \label{fig:filter_ab_fico}
\end{figure}

In this paragraph, we report counterparts to the results presented in \Cref{sec:experiments} using AdaBoost ensembles instead of random forests. Overall, similar trends are observed:
\begin{itemize}
    \item \Cref{fig:compu_time_ada} shows that large improvement in running time when using our proposed constraint-based formulation to retrieve separating samples compared to the objective-based one from \textsc{Fipe} is still observed with AdaBoost ensembles.
    \item \Cref{tab:pruning_ab} also outlines that the combination of learner generation and pruning in \textsc{Pace} is key to achieve strong performance. We also observe that learners in the compressed ensemble are most of the time generated by \textsc{Pace}, and were not present in the original ensemble. In contrast to random forest ensembles, the complete version of \textsc{Pace} hit the 20-hour time limit for the two largest configurations of the \textsc{Ionosphere} dataset.
    \item \Cref{fig:filter_ab_compas,fig:filter_ab_fico} show that with AdaBoost ensembles, pruning capabilities can also be largely enhanced by a fine control on the faithfulness guarantees. Although more gradual than for random forests, pruning capabilities are still significantly improved when $\conflevel > 0$ or $\plauslevel > 0$.
\end{itemize}

\begin{table}[hbtp]
    \centering
    \caption{Statistics of \textsc{Pace} and its ablated variants for AdaBoost ensembles with $n_{\text{est}}~\in~\{25,50,75,100\}$ trees of depth $d=2$ and faithfulness parameters $(\conflevel,\plauslevel)=(0.1,0.1)$. Values include the size $S$ of the ensemble resulting from \textsc{Pace} and its ablated variants, the percentage $P$ of learners that have been newly generated in this ensemble, and the running time $T$ in seconds. The notation ``\texttt{tlim}'' indicates that the 20-hours time budget has been exceeded.}
    \label{tab:pruning_ab}
    \setlength{\tabcolsep}{6pt}
    \resizebox{\textwidth}{!}{%
    \begin{tabular}{ll|rrr|rrr|rrr|rrr}
      \toprule
      & & \multicolumn{3}{c|}{$n_{\text{est}}=25$}  & \multicolumn{3}{c|}{$n_{\text{est}}=50$}  & \multicolumn{3}{c|}{$n_{\text{est}}=75$}  & \multicolumn{3}{c}{$n_{\text{est}}=100$} \\
      \textbf{Dataset} & \textbf{Method}  & $S$ & $P$ & $T$ & $S$ & $P$ & $T$ & $S$ & $P$ & $T$ & $S$ & $P$ & $T$ \\
      \midrule
      \rowcolor{gray!15} \textsc{Cancer} & Generate only & 25 & 100\% & 245 & 43 & 100\% & 518 & 70 & 100\% & 806 & 73 & 100\% & 830 \\
      \rowcolor{gray!15} & Pruning only & 17 & -- & 10 & 19 & -- & 19 & 19 & -- & 36 & 19 & -- & 42 \\
      \rowcolor{gray!15} & \textsc{Pace} & 16 & 100\% & 1,589 & 8 & 89\% & 22,726 & 9 & 84\% & 16,318 & 9 & 71\% & 1,134 \\
      \textsc{Compas}  & Generate only & 14 & 100\% & 23 & 16 & 100\% & 44 & 23 & 100\% & 72 & 14 & 100\% & 64 \\
      & Pruning only & 10 & -- & 3 & 8 & -- & 6 & 8 & -- & 9 & 9 & -- & 9 \\
      & \textsc{Pace} & 3 & 30\% & 167 & 5 & 75\% & 71 & 8 & 100\% & 93 & 4 & 100\% & 93 \\
      \rowcolor{gray!15} \textsc{Diabetes} & Generate only & 24 & 100\% & 220 & 35 & 100\% & 285 & 49 & 100\% & 569 & 57 & 100\% & 660\\
      \rowcolor{gray!15} & Pruning only & 13 & -- & 10 & 14 & -- & 19 & 12 & -- & 10 & 12 & -- & 23 \\
      \rowcolor{gray!15} & \textsc{Pace} & 8 & 100\% & 2,091 & 8 & 93\% & 1,966 & 8 & 52\% & 6,545 & 8 & 54\% & 4,791 \\
      \textsc{Elec2}  & Generate only & 21 & 100\% & 11 & 37 & 100\% & 260 & 47 & 100\% & 330 & 43 & 100\% & 350 \\
      & Pruning only & 12 & -- & 10 & 16 & -- & 17 & 14 & -- & 23 & 13 & -- & 25 \\
      & \textsc{Pace} & 9 & 92\% & 697 & 10 & 100\% & 1,201 & 10 & 93\% & 1,000 & 8 & 93\% & 695 \\
      \rowcolor{gray!15} \textsc{Fico} & Generate only & 20 & 100\% & 170 & 33 & 100\% & 254 & 28 & 100\% & 188 & 30 & 100\% & 164 \\
      \rowcolor{gray!15} & Pruning only & 11 & -- & 12 & 11 & -- & 12 & 12 & -- & 16 & 22 & -- & 29 \\
      \rowcolor{gray!15} & \textsc{Pace} & 8 & 91\% & 637 & 8 & 82\% & 671 & 8 & 100\% & 698 & 8 & 55\% & 1,429 \\
      \textsc{House-16H}  & Generate only & 25 & 100\% & 11 & 35 & 100\% & 312 & 57 & 100\% & 808 & 57 & 100\% & 809 \\
      & Pruning only & 13 & -- & 20 & 13 & -- & 167 & 13 & -- & 380 & 57 & -- & 654 \\
      & \textsc{Pace} & 1 & 100\% & 534 & 1 & 100\% & 833 & 1 & 100\% & 836 & 1 & 100\% & 830 \\
      \rowcolor{gray!15} \textsc{Htru2} & Generate only & 6 & 100\% & 17 & 20 & 100\% & 129 & 1 & 100\% & 8 & 21 & 100\% & 264 \\
      \rowcolor{gray!15} & Pruning only & 9 & -- & 1 & 8 & -- & 15 & 8 & -- & 20 & 3 & -- & 19 \\
      \rowcolor{gray!15} & \textsc{Pace} & 1 & 78\% & 560 & 7 & 62\% & 457 & 5 & 80\% & 652 & 3 & 100\% & 127 \\
      \textsc{Ionosphere}  & Generate only & 25 & 100\% & 312 & 49 & 100\% & 597 & 73 & 100\% & 1,065 & 96 & 100\% & 1,474 \\
      & Pruning only & 8 & -- & 18 & 12 & -- & 85 & 8 & -- & 114 & 12 & -- & 296 \\
      & \textsc{Pace} & 7 & 25\% & 7,478 & 10 & 48\% & 8,254 & \texttt{tlim} & \texttt{tlim} & \texttt{tlim} & \texttt{tlim} & \texttt{tlim} & \texttt{tlim} \\
      \rowcolor{gray!15} \textsc{Pol} & Generate only & 10 & 100\% & 7 & 16 & 100\% & 28 & 24 & 100\% & 80 & 25 & 100\% & 57 \\
      \rowcolor{gray!15} & Pruning only & 7 & -- & 1 & 8 & -- & 4 & 10 & -- & 4 & 14 & -- & 5 \\
      \rowcolor{gray!15} & \textsc{Pace} & 4 & 86\% & 214 & 6 & 100\% & 688 & 4 & 90\% & 2,287 & 4 & 100\% & 17,197 \\
      \textsc{Seeds}  & Generate only & 24 & 100\% & 203 & 42 & 100\% & 407 & 46 & 100\% & 386 & 47 & 100\% & 451 \\
      & Pruning only & 11 & -- & 16 & 13 & -- & 24 & 13 & -- & 12 & 12 & -- & 11 \\
      & \textsc{Pace} & 1 & 73\% & 631 & 12 & 92\% & 852 & 1 & 92\% & 1,166 & 1 & 100\% & 1,672 \\
      \rowcolor{gray!15} \textsc{Spambase} & Generate only & 25 & 100\% & 79 & 38 & 100\% & 493 & 46 & 100\% & 498 & 55 & 100\% & 1,018 \\
      \rowcolor{gray!15} & Pruning only & 17 & -- & 2 & 19 & -- & 6 & 13 & -- & 50 & 11 & -- & 27 \\
      \rowcolor{gray!15} & \textsc{Pace} & 8 & 94\% & 20,654 & 10 & 84\% & 64,950 & 10 & 44\% & 12,184 & 8 & 32\% & 11,829 \\
      \bottomrule
    \end{tabular}
    }
\end{table}

\FloatBarrier

\subsection{Comparison with a State-of-the-Art Pruning Method}
\label{app:compl_fipel0}

In this paragraph, we compare \textsc{Pace} with the state-of-the-art pruning method \textsc{Fipe} proposed by \citet{emine2025free}. The latter can be viewed as a special case of \textsc{Pace} in which no new learners are generated, and faithfulness is enforced globally (i.e., $\conflevel=0$ and $\plauslevel=0$). For a fair comparison, we evaluate \textsc{Pace} under the same faithfulness setting.
\Cref{tab:results_d1_n50} reports compression performance and running times for both methods. We consider the $\ell_0$-norm version of \textsc{Fipe}, which, within our 20-hour time limit, scales only to random forest and AdaBoost ensembles with $n_{\text{est}}=50$ trees of depth $d=1$. While the $\ell_1$-norm version of \textsc{Fipe} scales to larger models, it would have led to weaker pruning performance. Even in this small-scale regime where pruning opportunities are limited, \textsc{Pace} consistently achieves equal or better performance than \textsc{Fipe}:
\begin{itemize}
    \item On random forests, \textsc{Pace} outperforms \textsc{Fipe} on 6 out of the 11 datasets listed in \Cref{app:datasets}, and matches its performance on the remaining ones. The largest improvements are observed on \textsc{Htru2}, where \textsc{Pace} produces an ensemble that is 50\% smaller, and on \textsc{House-16H} and \textsc{Ionosphere}, with reductions of 36\% and 33\%, respectively.
    \item The gains are even more pronounced for AdaBoost ensembles. \textsc{Pace} outperforms \textsc{Fipe} on 8 of the 11 datasets listed in \Cref{app:datasets}, with particularly large margins on \textsc{Diabetes}, \textsc{Elec2}, \textsc{Pol}, and \textsc{Htru2}. On these datasets, \textsc{Pace} reduces the ensemble to a single tree, corresponding to improvements of 96\%, 94\%, 90\%, and 80\% over \textsc{Fipe}, respectively. The three datasets where both methods tie are also those in which no pruning gains are observed with random forests, suggesting limited redundancy at this scale.
\end{itemize}
Although \textsc{Pace} yields higher solution times, this experiment highlights the benefit of its two-stage process with both active learner generation and pruning. This combination is particularly beneficial for achieving good pruning performance in low-redundancy regimes where the initial ensemble cannot be pruned to a large extent by \textsc{Fipe}.

\begin{table}[hbtp]
  \centering
  \caption{Compression results for random forest and AdaBoost ensembles with $n_{\text{est}} = 50$ trees of depth $d=1$. Values include the size $S$ of the ensemble resulting from \textsc{Pace}, the percentage $P$ of learners that have been newly generated in this ensemble, and the overall running time $T$ in seconds.}
  \label{tab:results_d1_n50}
  \renewcommand{\arraystretch}{1.2}
  \small
  \begin{tabular}{ll|rrr|rrr}
    \toprule
    & & \multicolumn{3}{c|}{\textbf{Random forest}} & \multicolumn{3}{c}{\textbf{AdaBoost}} \\
    \textbf{Dataset} & \textbf{Method} & $S$ & $P$ & $T$ &  $S$ & $P$ & $T$ \\
    \midrule
    \rowcolor{gray!15}\textsc{Cancer} & \textsc{Pace} & 10 & 0\% & 21 & 20 & 0\% & 3 \\
    \rowcolor{gray!15} & \textsc{Fipe} & 10 & -- & 1 & 20 & -- & 9 \\
    \textsc{Compas} & \textsc{Pace} & 9 & 0\% & 16 & 13 & 0\% & 5 \\
    & \textsc{Fipe} & 9 & -- & 2 & 13 & -- & 6 \\
    \rowcolor{gray!15}\textsc{Elec2} & \textsc{Pace} & 5 & 18\% & 6 & 1 & 100\% & 1 \\
    \rowcolor{gray!15} & \textsc{Fipe} & 6 & -- & 2 & 18 & -- & 7 \\
    \textsc{Fico} & \textsc{Pace} & 10 & 0\% & 159 & 16 & 0\% & 5 \\
    & \textsc{Fipe} & 10 & -- & 3 & 16 & -- & 29 \\
    \rowcolor{gray!15}\textsc{Htru2} & \textsc{Pace} & 3 & 33\% & 3 & 1 & 100\% & 1 \\
    \rowcolor{gray!15} & \textsc{Fipe} & 6 & -- & 1 & 5 & -- & 1 \\
    \textsc{House-16H} & \textsc{Pace} & 9 & 11\% & 1,189 & 13 & 8\% & 2,370 \\
    & \textsc{Fipe} & 14 & -- & 13 & 31 & -- & 301 \\
    \rowcolor{gray!15}\textsc{Ionosphere} & \textsc{Pace} & 12 & 8\% & 9 & 13 & 8\% & 1 \\
    \rowcolor{gray!15} & \textsc{Fipe} & 18 & -- & 29 & 21 & -- & 12 \\
    \textsc{Diabetes} & \textsc{Pace} & 17 & 0\% & 4 & 1 & 100\% & 1 \\
    & \textsc{Fipe} & 17 & -- & 6 & 23 & -- & 9 \\
    \rowcolor{gray!15}\textsc{Pol} & \textsc{Pace} & 16 & 6\% & 3 & 1 & 100\% & 1 \\
    \rowcolor{gray!15} & \textsc{Fipe} & 20 & -- & 13 & 10 & -- & 1 \\
    \textsc{Seeds} & \textsc{Pace} & 15 & 0\% & 9 & 4 & 25\% & 1 \\
    & \textsc{Fipe} & 15 & -- & 10 & 7 & -- & 1 \\
    \rowcolor{gray!15}\textsc{Spambase} & \textsc{Pace} & 17 & 6\% & 690 & 20 & 5\% & 10 \\
    \rowcolor{gray!15} & \textsc{Fipe} & 20 & -- & 56 & 25 & -- & 65 \\
    \bottomrule
  \end{tabular}
\end{table}

\FloatBarrier

\subsection{Finer Analysis on the Generation of Separating Samples}
\label{app:compl_oracle}

\begin{figure}[hbtp]
    \centering

    \begin{subfigure}[t]{0.49\textwidth}
        \centering
        \includegraphics[width=\textwidth]{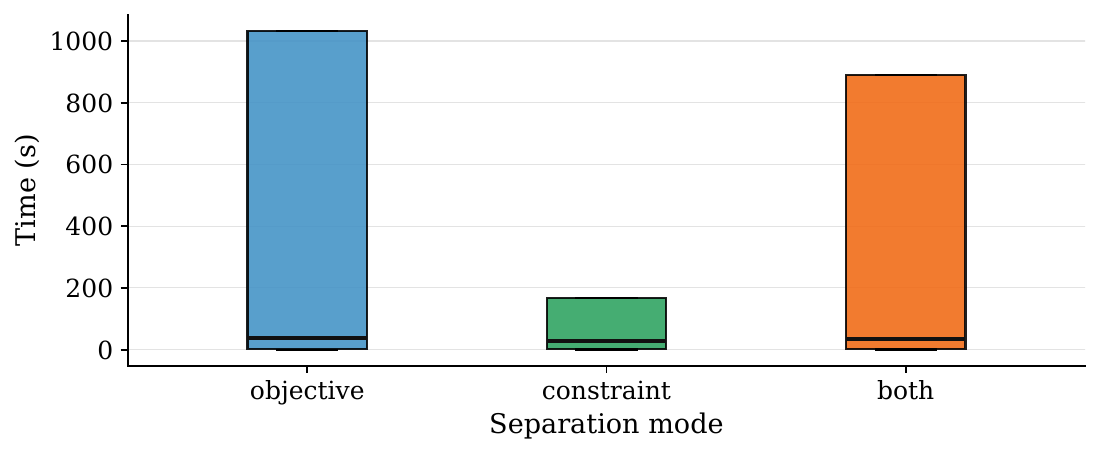}
    \end{subfigure}\hfill
    \begin{subfigure}[t]{0.49\textwidth}
        \centering
        \includegraphics[width=\textwidth]{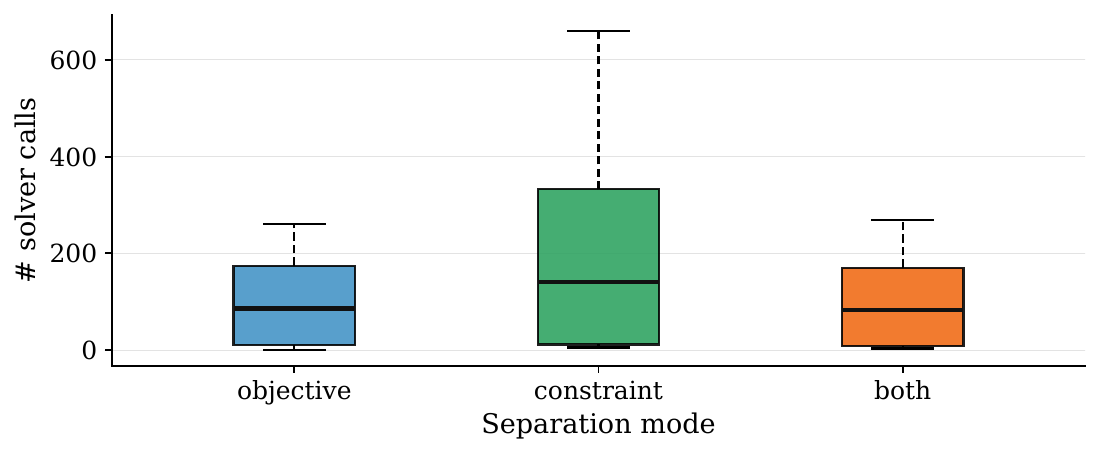}
    \end{subfigure}
    \caption{Overall computational time (left) and number of separation problems solved (right) until faithfulness can be guaranteed globally with Problem \eqref{prob:l1} using a random forest ensemble with $n_{\text{est}}=100$ trees of depth $d=3$. Results aggregate all the 11 datasets listed in \Cref{app:datasets}.}
    \label{fig:oracle_mode_rf}
\end{figure}

\begin{figure}[hbtp]
    \centering

    \begin{subfigure}[t]{0.49\textwidth}
        \centering
        \includegraphics[width=\textwidth]{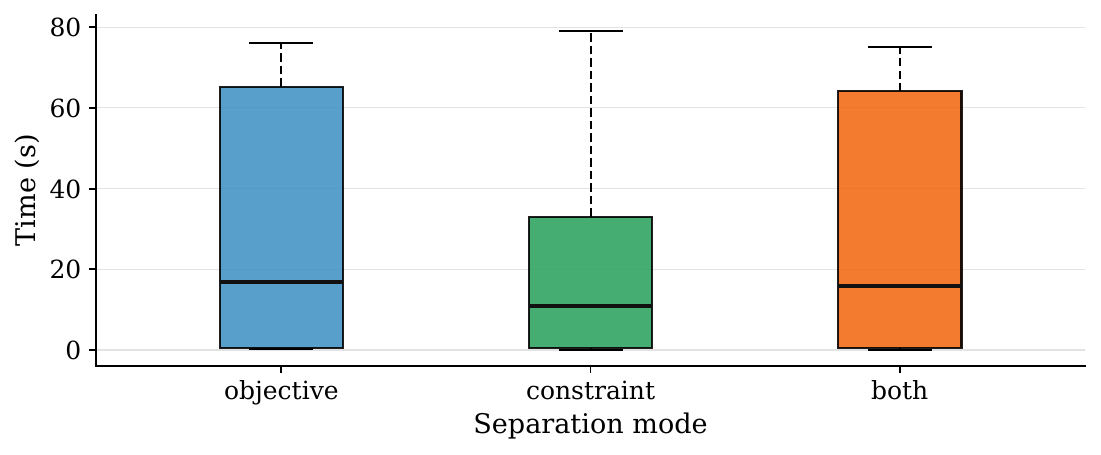}
    \end{subfigure}\hfill
    \begin{subfigure}[t]{0.49\textwidth}
        \centering
        \includegraphics[width=\textwidth]{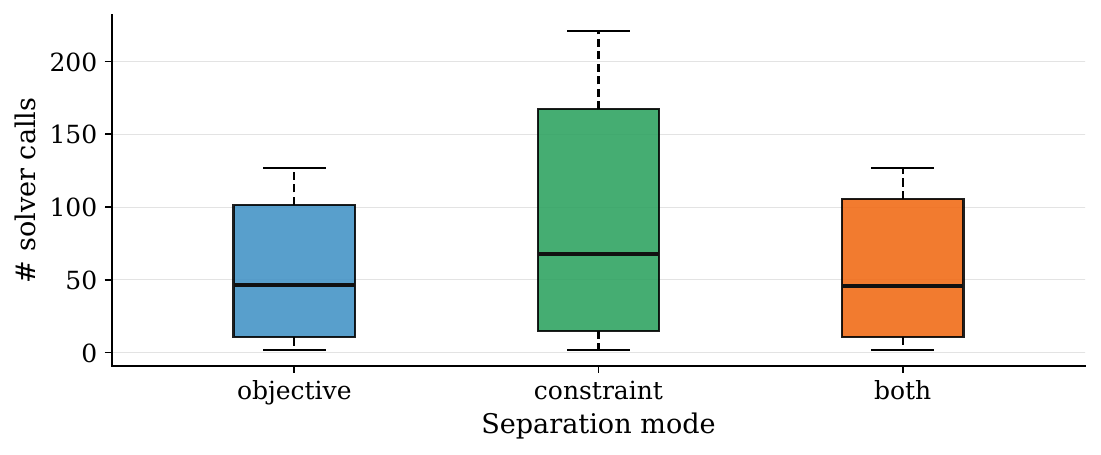}
    \end{subfigure}
    \caption{Overall computational time (left) and number of separation problems solved (right) until faithfulness can be guaranteed globally with Problem \eqref{prob:l1} using a AdaBoost ensemble with $n_{\text{est}}=100$ trees of depth $d=3$. Results aggregate all the 11 datasets listed in \Cref{app:datasets}.}
    \label{fig:oracle_mode_ab}
\end{figure}

In this paragraph, we further analyze the constraint-based formulation~\eqref{oracleconstr:separation_constraint} used in \textsc{Pace} to retrieve separating samples. We compare it to an objective-based separation problem
\begin{equation}
    \left\{
    \begin{array}{lll}
        \textstyle \max_{\sample{} \in \samplespace} & \textstyle\sum_{\learner{} \in \learners\phantom{{}^o}} \ 
        \learnerweight{\learner{}} \cdot (r_{\learner{}\predict{}} -r_{\learner{}\initpredict}) \\
        \text{s.t.}     &\textstyle\sum_{\learner{} \in \initlearners} \ 
        \initlearnerweight{\learner{}} \cdot (r_{\learner{}\initpredict} -r_{\learner{}\tilde{\predict{}}}) &> 0 \quad \forall \tilde{\predict{}} \neq \initpredict
    \end{array}
    \right.
\end{equation}
which corresponds to prior separation formulations proposed in the literature \citep{emine2025free}. 
These formulations opt for different trade-offs:
\begin{itemize}
    \item Objective-based separation seeks maximal disagreement in prediction, typically yielding fewer but higher-quality separating samples, at the expense of higher computational cost.
    \item Constraint-based separation seeks any separating sample, often resulting in more separating samples generated, but at a lower computational cost since only feasibility is required.
\end{itemize}
This trade-off is investigated in \Cref{fig:oracle_mode_rf,fig:oracle_mode_ab} where we report the overall solving time and number of samples generated when solving Problem \eqref{prob:l1} until faithfulness can be certified globally over $\samplespace = \R^{\sampledim}$, setting $\learners=\initlearners$ and initializing $\initsamplespace$ with the dataset used to train the initial ensemble $(\initlearners, \initlearnerweights)$.
Results show that constraint-based separation consistently achieves the lowest computational time, although it requires generating more separating samples than its objective-based counterpart. Combining these approaches by enforcing separation with both an objective and a constraint does not yield meaningful improvements in runtime, even if the number of generated samples is reduced compared to the constraint-based separation. The constraint-based approach appears to be the best choice overall.

\FloatBarrier

\subsection{Finer Analysis on the Generation of Improving Learners}
\label{app:compl_cg}

\begin{figure}[hbtp]
    \centering
    \begin{subfigure}{\textwidth}
        \centering
        \includegraphics[clip, trim=0.1cm 0.1cm 0.1cm 0.1cm,
                         width=0.75\textwidth]{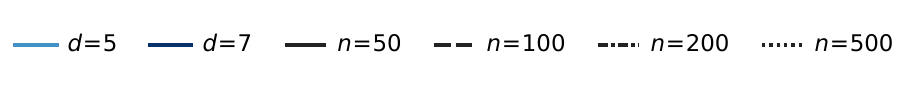}
    \end{subfigure}
    \begin{subfigure}[t]{0.49\textwidth}
        \centering
        \includegraphics[width=\textwidth]{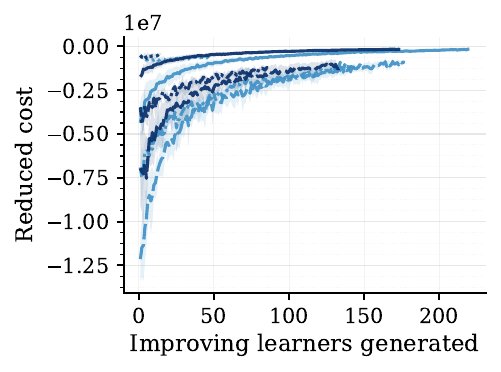}
    \end{subfigure}\hfill
    \begin{subfigure}[t]{0.49\textwidth}
        \centering
        \includegraphics[width=\textwidth]{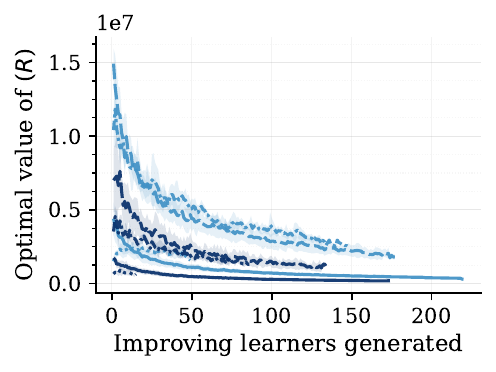}
    \end{subfigure}
    \begin{subfigure}[t]{0.49\textwidth}
        \centering
        \includegraphics[width=\textwidth]{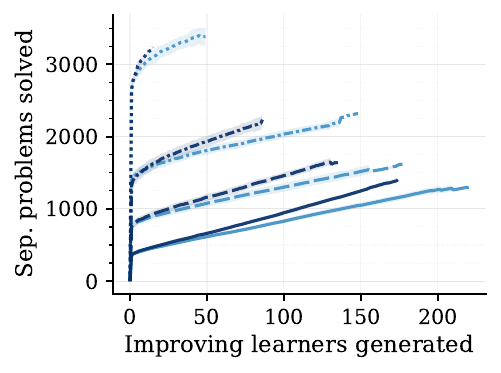}
    \end{subfigure}\hfill
    \begin{subfigure}[t]{0.49\textwidth}
        \centering
        \includegraphics[width=\textwidth]{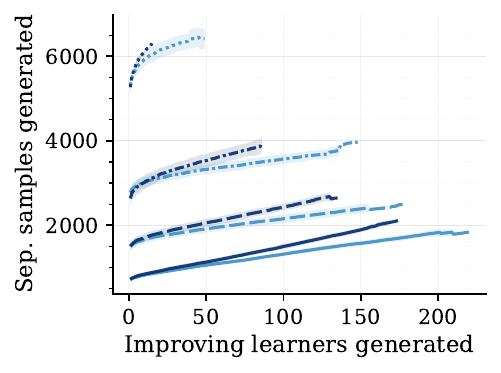}
    \end{subfigure}
    \caption{Statistics of the improving learner generation phase of \textsc{Pace} for random forest ensembles on the \textsc{Ionosphere} dataset, with $n_{\text{est}} \in \{50, 100, 200, 500\}$ trees of depth $d \in \{5, 7\}$. Each panel shows the mean over five independent runs of the experiment and shading indicates standard deviation.}
    \label{fig:convergence_rf_ionosphere}
\end{figure}

\begin{figure}[hbtp]
    \centering
    \begin{subfigure}{\textwidth}
        \centering
        \includegraphics[clip, trim=0.1cm 0.1cm 0.1cm 0.1cm,
                         width=0.75\textwidth]{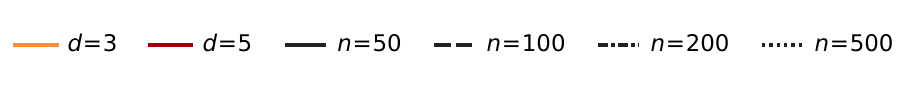}
    \end{subfigure}
    \vspace{-0.4em}
    \begin{subfigure}[t]{0.49\textwidth}
        \centering
        \includegraphics[width=\textwidth]{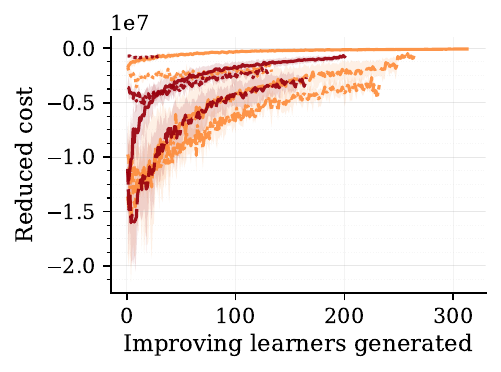}
    \end{subfigure}\hfill
    \begin{subfigure}[t]{0.49\textwidth}
        \centering
        \includegraphics[width=\textwidth]{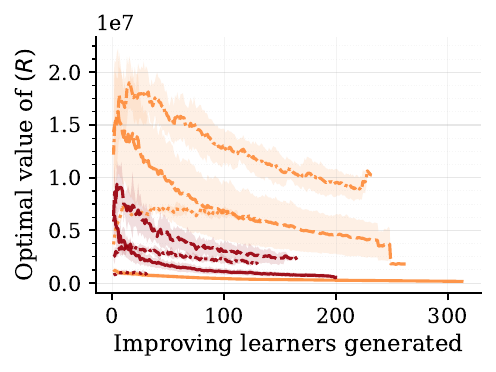}
    \end{subfigure}
    \begin{subfigure}[t]{0.49\textwidth}
        \centering
        \includegraphics[width=\textwidth]{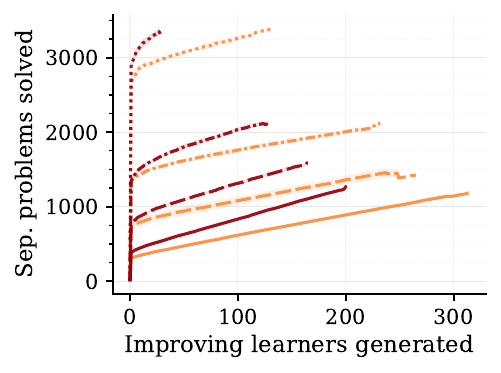}
    \end{subfigure}\hfill
    \begin{subfigure}[t]{0.49\textwidth}
        \centering
        \includegraphics[width=\textwidth]{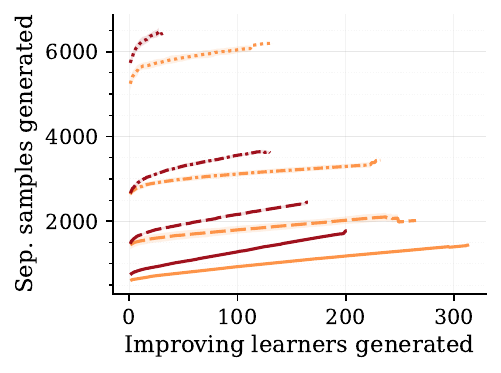}
    \end{subfigure}
    \caption{Statistics of the improving learner generation phase of \textsc{Pace} for AdaBoost ensembles on the \textsc{Ionosphere} dataset, with tree $n_{\text{est}} \in \{50, 100, 200, 500\}$ of depth $d \in \{5, 7\}$. Each panel shows the mean over five independent runs of the experiment and shading indicates standard deviation.}
    \label{fig:convergence_ab_ionosphere}
\end{figure}

In this paragraph, we further analyze the first phase where improving learners are generated by \textsc{Pace}. We conduct experiments on the \textsc{Ionosphere} dataset, compressing random forest and AdaBoost ensembles with $n_{\text{est}} \in \{50,100,200,500\}$ trees of depths $d \in \{5,7\}$. \Cref{fig:convergence_rf_ionosphere,fig:convergence_ab_ionosphere} report the evolution of the reduced cost in \Cref{prop:pricing}, the optimal value of Problem \eqref{prob:l1}, the cumulative number of separation problems solved, and the cumulative number of separating samples generated, as a function of the number of improving learners generated by \textsc{Pace}.

First, the top-left plots in \Cref{fig:convergence_rf_ionosphere,fig:convergence_ab_ionosphere} show that the reduced cost consistently converges toward zero across all tested configurations, indicating diminishing opportunities for improving the optimal value of Problem \eqref{prob:l1} as new improving learners are added. This trend is reflected in the top-right plots, where the optimal value of Problem \eqref{prob:l1} progressively decreases. Convergence appears slower for more complex ensembles with deeper, more numerous trees. In some cases, the reduced cost does not reach zero due to the 20-hour time limit, which interrupts the learner generation phase of \textsc{Pace}.

The bottom plots in \Cref{fig:convergence_rf_ionosphere,fig:convergence_ab_ionosphere} show that most separating samples are generated during the very first iteration of the learner generation phase in \textsc{Pace}. In subsequent iterations, both the number of separation problems solved and the number of newly generated samples per iteration stabilize, with roughly linear growth as the number of improving learners increases. This suggests that relevant separating samples to enforce faithfulness are largely captured when pruning the initial ensemble given to \textsc{Pace}. Subsequent iterations mainly perform incremental improvements with new learners incorporated into the ensemble, and only minor corrections to ensure faithfulness are performed.

